JOÃO EUGENIO MARYNOWSKI

# *AMBIENTE DE PLANEJAMENTO IPÊ*

Dissertação apresentada como requisito parcial à obtenção do grau de Mestre. Programa de Pós-Graduação em Informática, Setor de Ciências Exatas, Universidade Federal do Paraná.
Orientador: Prof. Dr. Marcos Castilho

CURITIBA

2004

Dedico esta dissertação de mestrado aos meus pais, Eugenio Marynowski e Marília de Jesus Marynowski.

# Agradecimentos

Agradeço primeiramente a Deus por me conceder sua preciosa companhia me amparando e me auxiliando nos momentos mais difíceis.

Agradeço à minha família, meu pai Eugenio, minha mãe Marília, minha irmã Márcia e minha sobrinha Rafaela, pelo apoio incondicional e compreensão em meus momentos de ausência, que reconheço não terem sido poucos. Lembro aqui da minha namorada Corina que tanto me ajudou nesses e em outros momentos difíceis, muito obrigado.

Agradeço ao meu orientador e amigo Marcos Alexandre Castilho pela confiança, direção, incentivo, conselho e apoio dados durante esse trabalho.

Ao meu amigo Fabiano Silva que foi o principal responsável para esse feito. Foi meu professor durante a minha graduação, fez uma excelente propaganda de sua área de pesquisa de modo tal que não consegui escapar e este foi o resultado. Muito obrigado por sua ajuda em todos os aspectos.

Aos professores Luiz Allan e Andre Guedes pela ótima contribuição e aos meus amigos do mestrado, Andreas, Araújo, Bona, Egon, Evandro, Gabriel, Rodrigo, Thiago, Tiago, e a todos que me auxiliaram e me proporcionaram um excelente ambiente de trabalho.

Ao Departamento de Informática da UFPR, especialmente ao professor e amigo Alexandre I. Direne por proporcionar o suporte institucional e pelo seu grande incentivo.

Um agradecimento especial a Jusefina e a Jusefa, por tantas horas de trabalho.

# SUMÁRIO

# Resumo

Neste trabalho investigamos sistemas que implementam algoritmos para o problema de planejamento em Inteligência Artificial, denominados planejadores, com especial atenção aos baseados no grafo de planos. Analisamos o problema de se comparar o desempenho dos diferentes algoritmos e propomos um ambiente para facilitar o desenvolvimento e análise de planejadores.

**Palavras chave:** Inteligência Artificial, Planejamento e Grafo de Planos.

# Abstract

In this work we investigate the systems that implements algorithms for the planning problem in Artificial Intelligence, called planners, with especial attention to the planners based on the plan graph. We analyze the problem of comparing the performance of the different algorithms and we propose an environment for the development and analysis of planners.

**Keywords:** Artificial Intelligence, Planning and Plangraph.

# Capítulo 1

# Introdução

Uma das habilidades mais importantes do ser humano é conseguir planejar ações que levam a ter um determinado objetivo atingido. Por exemplo, como pode-se chegar na universidade em 15 minutos para não se perder a reunião? Quais as ações que devemos executar para solucionar esse problema? Encontrar estas ações utilizando um computador é o objeto de estudo no *Planejamento em Inteligência Artificial*.

Neste contexto, um *planejador* é um sistema que recebe como entrada um estado inicial, um conjunto de ações possíveis, o estado final desejado e gera um *plano*. Por exemplo, o *planejador* recebe o estado inicial "estar em casa", diversas ações possíveis e o estado final desejado "estar na universidade em 15 minutos" e retorna um *plano* que pode ser a seqüência: "pegue o telefone", "ligue para um táxi", "entre no táxi" e "diga ao motorista: preciso estar na universidade em 10 minutos".

Planejamento é um dos temas mais antigos na Inteligência Artificial (IA). As primeiras abordagens datam dos anos 60 no contexto de um subproblema da área de prova automática de teoremas em lógica de primeira ordem [Gre69]. Essa abordagem apresentava diversos problemas, sendo ainda ineficiente ao resolver problemas de planejamento.

Em 1971 Fikes e Nilsson definiram uma linguagem simplificada para a descrição dos problemas. Conhecida como *STRIPS* [FN71], essa linguagem permitiu tratar o planejamento como um problema de busca em um espaço de estados. Apesar de ter simplificado significativamente o problema, a complexidade teórica para esta categoria está em *PSPACE-Completo* [Byl94] e é denominada planejamento clássico.

Até a primeira metade dos anos 90 os planejadores desenvolvidos ainda eram baseados em antigas técnicas de busca, o que era insuficiente para tratar problemas simples que geravam amplos espaços de busca. A partir de então, novas idéias surgiram. A primeira introduziu a busca no espaço de planos, onde se tentava construir um plano a partir de operadores de planos. A partir de um plano vazio, se construíam seqüências de ações que tentavam transformar o estado inicial no estado final. O principal representante desta técnica é o UCPOP [PW92].

A segunda idéia foi baseada nos novos e rápidos algoritmos que foram propostos para resolver problemas de satisfatibilidade (SAT) [KS92]. Passando-se por um processo de redução do problema de planejamento a um problema de se resolver uma instância SAT foi possível obter excelentes resultados com alguns problemas não triviais.

Finalmente, foi encontrada uma estrutura de dados que permitiu representar o espaço de busca de maneira mais compacta e conseqüentemente levou à implementação de um algoritmo sofisticado e eficiente que resolveu vários problemas difíceis. Esta estrutura de dados é o grafo de planos e o algoritmo o *GRAPHPLAN* [BF95].

Desde então foram apresentados muitos outros algoritmos baseados no grafo de planos. Também o planejamento foi relacionado com outras áreas de pesquisa, tais como a Programação Inteira [VBLN99], a Programação por Restrição [BC99], as Redes de Petri [SCK00], entre outras. A figura 1.1 ilustra as diferentes técnicas aplicadas hoje no planejamento, como elas se relacionam e indica referências aos trabalhos que propuseram as respectivas abordagens:

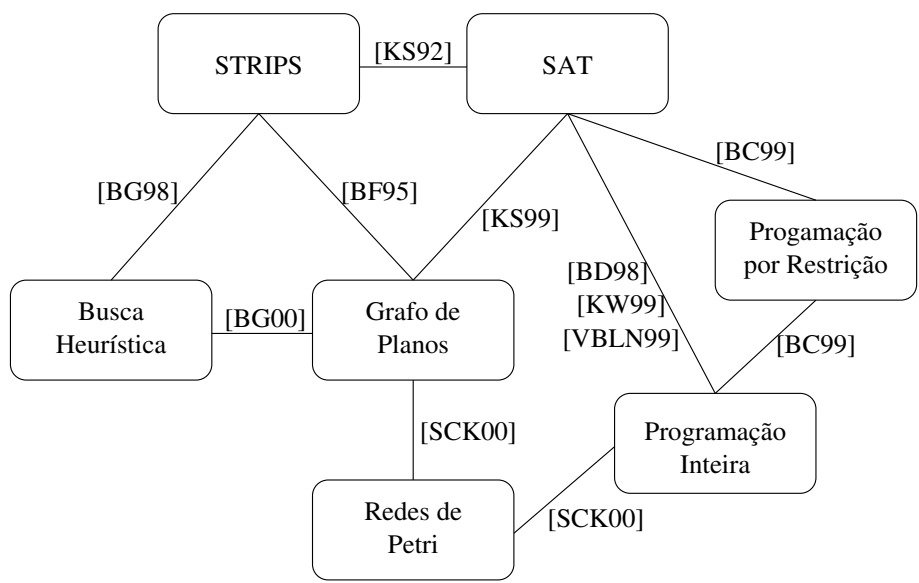

Figura 1.1: Diferentes técnicas para o planejamento.

Estes importantes avanços motivaram o resgate das antigas técnicas de busca, resultando em planejadores extremamente rápidos, tais como o *HSP* [BG98] e o *FF* [HN01], que obtiveram êxito ao definir ótimas funções heurísticas para guiar a busca. O *FF*, por exemplo, usa o grafo de planos como base para uma função heurística em um processo de subida de encosta e é considerado hoje o melhor sistema planejador.

A própria evolução da computação e das técnicas de programação permitiu um grande avanço no desenvolvimento de planejadores. Um fato que caracteriza fortemente essa evolução é a reimplementação do algoritmo *STRIPS* original que resultou em um planejador que resolve problemas difíceis [Lin01]. A figura 1.2 apresenta a hierarquia dos principais planejadores que surgiram de 1971 a 2000.

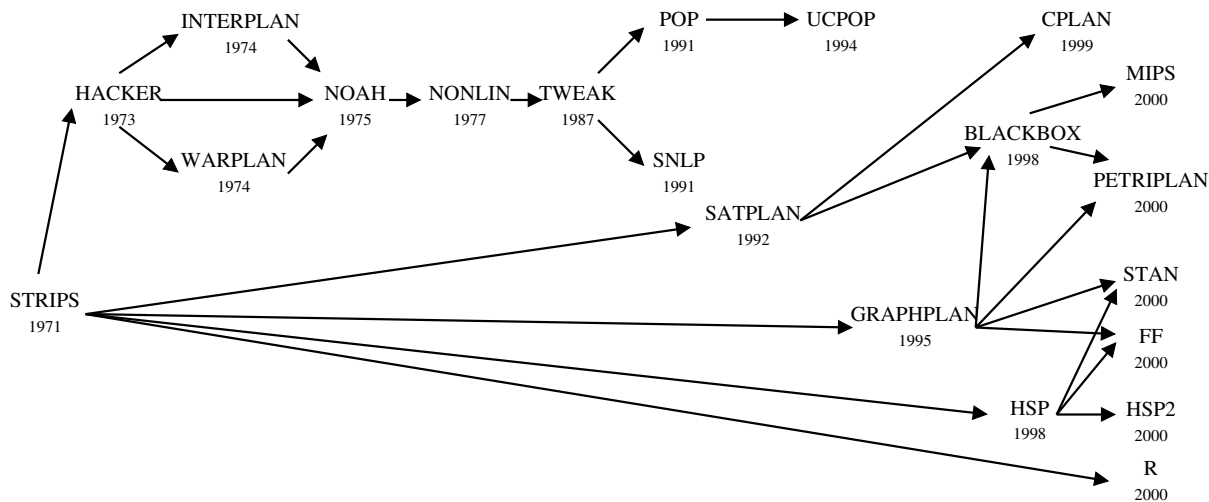

Figura 1.2: Hierarquia dos principais planejadores que surgiram de 1971 a 2000.

Um ponto interessante a ser observado é a dificuldade em se comparar e analisar estes diferentes planejadores. Isto não é tarefa fácil pois os planejadores são construídos em diferentes plataformas usando diferentes linguagens de programação, diferentes estruturas de dados e são baseados em diferentes linguagens de interface.

Para tentar resolver este problema a comunidade de planejamento criou a Competição Internacional de Planejadores, evento bienal cuja primeira versão aconteceu em 1998. Nesse ano foi definida uma linguagem unificada para descrever problemas de planejamento, denominada *PDDL* [McD98b]. Essa linguagem foi formada pela junção das linguagens *STRIPS* e *ADL* [Ped89] e é considerada hoje a linguagem de descrição padrão para a comunidade de planejamento. A escolha de uma linguagem de programação comum não foi considerada, mas a maioria dos planejadores estão escritos em C ou C++.

O ponto relevante a respeito das competições é que os planejadores são somente comparados observando os resultados finais obtidos com relação ao tempo gasto ou ao tamanho do plano, o que não é suficiente para permitir análises mais profundas. Ainda, ao observar somente os resultados finais torna-se difícil declarar claramente um ganhador ou que um algoritmo é melhor que os outros. Por exemplo, na competição de 1998, foi difícil apontar o vencedor [McD98a].

Na primeira competição foram avaliados o número de problemas resolvidos, o tempo total gasto para resolvê-los e o tamanho dos planos encontrados. A cada nova versão da competição são adicionados mais critérios para essa avaliação sempre em busca de melhor avaliar o desempenho dos planejadores. Neste sentido, os planejadores são considerados como entidades que resolvem problemas a partir de uma descrição comum, sem considerar como são resolvidos.

Em nossa opinião isto é insuficiente como análise. O que está sendo feito é analisar programas completos e isto empobrece a análise. Na nossa visão, planejadores são algoritmos sobre estruturas de dados. Nossa tentativa é analisar os planejadores recentes em termos de suas estruturas de dados (representações) e do algoritmo planejador (resolvedor). Por exemplo o planejador *BLACKBOX* [KS99] trabalha em duas fases, na primeira é construída a estrutura de dados e na segunda obtém-se a solução via um resolvedor SAT. Melhorando-se a representação, melhora-se o algoritmo final.

Neste trabalho propomos um modo complementar para comparar os planejadores em relação às competições. Propomos um ambiente comum baseado em conceitos orientados à objetos para construir planejadores: o Ambiente de Planejamento Ipê (*IPE - Ipê Planning Environment*). Sua característica principal é a separação clara dos processos envolvidos em um planejador: a descrição dos problemas, a estrutura de representação e o resolvedor.

Temos como meta disponibilizar um ambiente geral para planejamento permitindo o desenvolvimento e a integração de formas diferentes de representação, diferentes algoritmos e modos diferentes de descrever os problemas a serem resolvidos. A idéia é permitir a implementação de sistemas que interpretem problemas descritos em *PDDL*, *STRIPS*, *ADL* ou possivelmente outras linguagens, gerando representações como grafo de planos, redes de Petri, instâncias *SAT*, entre outras. Também deve possibilitar a implementação de diferentes resolvedores que utilizarão essas representações. Queremos dar liberdade para o programador escolher a combinação desejada de linguagem de descrição, representação e algoritmo resolvedor.

Outra motivação é educacional. O *IPE* pode servir como uma ótima ferramenta para cursos de Inteligência Artificial e Planejamento para a graduação e/ou pós-graduação. Algumas estruturas básicas já estão finalizadas, por exemplo: o interpretador de descrições em *PDDL*, as representações grafo de planos e redes de Petri, os resolvedores de busca exaustiva e alcançabilidade, bem como a interface. Os estudantes podem, portanto, implementar outros algoritmos e representações conhecidas, desenvolver novos algoritmos e representações ou podem simular uma competição inteira comparando e analisando melhor as diferenças entre os planejadores.

O texto está estruturado como segue. No capítulo 2 é feita uma abordagem geral do problema de planejamento em IA apresentando em detalhes o funcionamento dos planejadores, destacando a representação grafo de planos e os principais planejadores nele baseados. No capítulo 3 apresentamos o *IPE*, destacando sua arquitetura e uso. No capítulo 4 é feita uma análise envolvendo três planejadores implementados no *IPE*. No capítulo 5 são apresentadas algumas conclusões e propostas para trabalhos futuros.

# Capítulo 2

# O problema de planejamento em IA

Neste capítulo apresentaremos os fundamentos da área de planejamento em Inteligência Artificial, particularmente no que se refere ao chamado planejamento clássico ou planejamento baseado na linguagem *STRIPS*.

Mostraremos a definição formal do problema de planejamento, a linguagem padrão para descrição de domínios e problemas, as estruturas de representação mais importantes bem como os principais algoritmos empregados. Nosso objetivo é apresentar a linguagem *PDDL*, a representação grafo de planos e os principais algoritmos nela baseados.

## 2.1 Fundamentos

Formalmente, um *problema de planejamento* $\mathcal{P} = <\mathcal{O}, \mathcal{I}, \mathcal{G}>$ é uma tupla onde $\mathcal{O}$ é o conjunto de ações, $\mathcal{I}$ é o estado inicial, e $\mathcal{G}$ é o estado final. Um plano solução para $\mathcal{P}$ é uma seqüência de ações de $\mathcal{O}$ que transformam $\mathcal{I}$ em $\mathcal{G}$.

Considere por exemplo um comerciante que deseja transportar um conjunto de pacotes do depósito para suas duas lojas, para reabastecer seus estoques locais. Esse problema pode ser representado como um conjunto de ações em $\mathcal{O}$ descritas como: "carregar pacote $p$ no caminhão $c$", "dirigir o caminhão $c$ da localidade $l_1$ para a localidade $l_2$", "descarregar o pacote $p$ do caminhão $c$".

O estado inicial, contido em $\mathcal{I}$, representa o atual estoque de cada loja e $\mathcal{G}$ é a situação desejada pelo comerciante. Uma solução para esse problema, considerando que o caminhão esteja no depósito, é uma seqüência de ações semelhante à:

1. carregar o caminhão com os pacotes, $p_1, p_2, ..., p_n$

2. dirigir o caminhão do depósito para a *loja*1.

3. descarregar os pacotes $p_1, p_2, ..., p_i$ na *loja*1.

4. dirigir o caminhão da *loja*1 para a *loja*2.

5. descarregar os pacotes $p_{i+1}$, $p_{i+2}$,..., $p_n$ na *loja2*.

Existem diversas maneiras de se representar as informações que definem um problema de planejamento no computador. As primeiras abordagens datam dos anos 60, quando se usava uma descrição baseada em Lógica de Primeira Ordem (*LPO*). Isto resultava em diversos problemas, o principal deles era o problema da Persistência [MH69], além, evidentemente, dos problemas relativos ao processo de Prova Automática de Teoremas em *LPO* [Gre69].

Em 1971, Fikes e Nilsson propuseram um formalismo baseado não em lógica, mas em um processo de transformações controladas da descrição do estado usando um engenhoso mecanismo de listas. Este mecanismo permitiu definir o problema de planejamento como uma busca em um espaço de estados. O algoritmo e a linguagem de representação por eles apresentados ficaram conhecidos como *STRIPS* [FN71].

A idéia da linguagem *STRIPS* é representar os estados do mundo através de conjunções de literais instanciados livres de funções, ou seja, predicados aplicados sobre constantes (*LPO*) também chamados proposições. Na descrição dos estados assume-se que todas as fórmulas atômicas não explicitamente listadas são falsas, o que é chamado de "Hipótese do Mundo Fechado".

Cada ação $o$ é definida pela tupla: $o = (pre(o), add(o), del(o))$, onde $pre(o)$ é a lista de pré-condições, $add(o)$ é a lista de efeitos adicionados, e $del(o)$ é a lista de efeitos removidos. A lista de pré-condições é uma conjunção de literais positivos que devem ser verdadeiros para que a ação possa ser aplicada. A lista de efeitos adicionados é uma conjunção de literais positivos que serão incluídos no próximo estado e a lista de efeitos removidos é uma conjunção de literais que serão removidos no próximo estado.

O resultado da aplicação de uma ação $o$ em um estado $S$ é definido como a adição das proposições da lista $add(o)$ e a remoção das proposições da lista $del(o)$ somente se as proposições da lista $pre(o)$ existirem no estado $S$.

O conjunto $\mathcal{O}$, que define as ações possíveis segundo o problema, é dado pela instanciação de todas as ações com todos os objetos, podendo gerar algumas ações que não serão usadas e algumas que não serão válidas, mas todas devem ser instanciadas para serem avaliadas.

A instanciação é o processo pelo qual as variáveis que são utilizadas na definição das ações são substituídas pelos objetos disponíveis no problema a ser resolvido. Essa definição será apresentada em mais detalhes na seção seguinte.

Em planejamento pode-se classificar um problema como sendo *relaxado*, onde não são considerados os efeitos de remoção das ações. Assim, um *problema de planejamento relaxado* $\mathcal{P}'$ de um problema de planejamento $\mathcal{P}$, é formalmente definido como: $\mathcal{P}' = <\mathcal{O}', \mathcal{I}, \mathcal{G}>$, onde:

$$\mathcal{O}' = \{o' = (pre(o), add(o), \emptyset) | o \in \mathcal{O}\}$$

Em outras palavras, o problema de planejamento relaxado é obtido ignorando-se as listas de efeitos removidos das ações. Esse conceito será abordado em mais detalhes na seção 2.3.1.

O formalismo baseado em regras apresentado pela linguagem *STRIPS* permitiu a implementação efetiva de planejadores, mesmo este sendo um formalismo extremamente limitado, pois permite representar apenas uma classe restrita de problemas, conhecida como planejamento clássico. Nesta categoria os problemas são sempre determinísticos e o mundo é totalmente conhecido. As ações não tem duração e não são considerados recursos como tempo, distância e consumo. O planejamento não clássico permite envolver aspectos complexos como replanejamento, universos dinâmicos e quantificadores universais. Neste contexto, estaremos interessados somente no planejamento clássico.

Nas três próximas seções apresentaremos um planejador visto a partir do que consideramos ser suas três partes fundamentais: a linguagem de representação, a estrutura de dados e os algoritmos em si.

## 2.2 A linguagem *PDDL*

A linguagem *PDDL* - "*The Planning Domain Definition Language*" [McD98b] foi desenvolvida em 1998 com o objetivo de unificar a definição de problemas para sistemas de planejamento, permitindo assim compará-los mais facilmente a partir de um mesmo conjunto de problemas.

Essa linguagem é uma combinação das linguagens *STRIPS* e *ADL* e é capaz de representar: efeitos condicionais; quantificadores universais; universos dinâmicos, permitindo a criação e a destruição de objetos; axiomas de domínio; especificação de ações hierárquicas compostas de sub-ações e sub-metas e gerenciamento de múltiplos problemas em múltiplos domínios.

Um domínio é definido basicamente pela descrição das ações, ou seja, o conjunto $\mathcal{O}$ referenciando o problema de planejamento $\mathcal{P} = <\mathcal{O}, \mathcal{I}, \mathcal{G}>$. Os conjuntos $\mathcal{I}$ e $\mathcal{G}$ definem um problema particular para esse domínio. Assim podem haver vários problemas para cada domínio, ou seja, várias combinações de estados iniciais e finais. Por exemplo, segundo o nosso exemplo do comerciante, pode-se querer levar somente dois pacotes para a *loja*1, ou quatro para a *loja*2, caracterizando dois problemas distintos para o mesmo domínio.

Mais detalhadamente, a representação de um domínio em *PDDL* é dada por uma declaração contendo os seguintes itens:

- a definição do nome do domínio;

- o conjunto de características de representação necessárias para a definição do domínio;

- o conjunto de predicados presentes no domínio; e

- as ações que podem ser executadas no domínio, descritas por:

    - nome da ação;

    - parâmetros passados para a ação;

    - conjunção das pré-condições da ação; e

    - conjunção dos efeitos da ação.

A representação de um problema em *PDDL* é descrita por uma declaração contendo os seguintes itens:

- definição do nome do problema;

- domínio ao qual pertence;

- conjunto de objetos existentes no problema;

- estado inicial; e

- estado final;

A sintaxe da linguagem *PDDL* é simples mas apresenta um grande número de itens para possibilitar a representação de todas as funcionalidades por ela descritas. Como exemplo será aqui apresentado apenas um sub-conjunto de sua capacidade representativa. Um maior detalhamento da sintaxe e da semântica da linguagem pode ser encontrado em [McD98b].

Usaremos apenas três itens da sintaxe: as palavras iniciadas por ":" são rótulos reservados, as iniciadas por "?" são variáveis e as palavras depois do "-" são tipos. A parentização define o limite de determinada definição ou opção.

Descrevendo nosso exemplo do comerciante segundo a linguagem *PDDL*, temos a definição do domínio como:

```
(define (domain comerciante)
(:types caminhão - veículo
          pacote
          veículo - objeto
          localidade
          objeto - geral)

  (:predicates
                  (em ?obj - objeto ?loc - localidade)
                  (dentro ?pkg - pacote ?veh - veículo))

(:action carregar
   :parameters    (?pkg - pacote ?truck - caminhão ?loc - localidade)
```

```
    :precondition  (and (em ?truck ?loc) (em ?pkg ?loc))
    :effect        (and (not (em ?pkg ?loc)) (dentro ?pkg ?truck)))

(:action descarregar
  :parameters   (?pkg - pacote ?truck - caminhão ?loc - localidade)
  :precondition (and (em ?truck ?loc) (dentro ?pkg ?truck))
  :effect       (and (not (dentro ?pkg ?truck)) (em ?pkg ?loc)))

(:action dirigir
  :parameters (?truck - caminhão ?loc-from - localidade ?loc-to - localidade)
  :precondition (and (em ?truck ?loc-from) )
  :effect (and (not (em ?truck ?loc-from)) (em ?truck ?loc-to)))
)
```

A primeira linha contém o nome do domínio. Da segunda à sétima linha são descritos os tipos existentes e suas hierarquias. Da nona à décima primeira linha são descritos os predicados. Todos os predicados devem estar contidos nesse conjunto pois, além de formar os estados possíveis, são utilizados na descrição das ações.

A primeira ação descrita é *carregar*, esta ação possui três parâmetros: um pacote, um caminhão e uma localidade; duas pré-condições: o caminhão estar na localidade assim como o pacote; e dois efeitos, um de remoção: negando que o pacote está na localidade e um de adição: o pacote está dentro do caminhão.

A segunda ação descrita é *descarregar*; esta ação tem os mesmos parâmetros que a ação anterior, mudando apenas as pré-condições: o caminhão estar na localidade e o pacote estar dentro do caminhão; e os efeitos: negando que o pacote está dentro do caminhão e adicionando que o pacote está na localidade.

A terceira e última ação descrita é *dirigir*, que também possui três parâmetros: um caminhão, uma localidade origem e uma localidade destino; uma pré-condição: o caminhão estar na localidade origem; e dois efeitos, um de remoção: negando que o caminhão está na localidade origem e um de adição: o caminhão está na localidade destino.

O problema, denominado aqui *comerciante-1*, é descrito como:

```
(define (problem comerciante-1)
  (:domain comerciante)
  (:objects pacote1 pacote2 - pacote
            caminhão1 - caminhão
            loja1 loja2 depósito - localidade)
  (:init
        (em caminhão1 depósito)
        (em pacote1 loja1)
        (em pacote2 loja1))
  (:goal (and (em pacote2 loja2)
              (em pacote2 loja2) )))
```

De forma análoga ao domínio, temos na primeira linha a definição do nome do problema, *comerciante-1*. Na segunda linha é definido o domínio do qual este problema faz parte, no nosso caso *comerciante*. Na terceira linha são definidos os objetos existentes: *pacote*1 e *pacote*2 do tipo *pacote*; *caminhão*1 do tipo *caminhão*; e *loja*1, *loja*2 e *depósito* como sendo do tipo *localidade*.

O rótulo reservado ":init", define o início da descrição do estado inicial formado pelas proposições: *caminhão*1 no *depósito*; *pacote*1 na *loja*1; *pacote*2 na *loja*1. De forma análoga, o rótulo reservado ":goal" define o início da descrição do estado final formado pelas proposições: *pacote*1 na *loja*2; *pacote*2 na *loja*2.

Como foi apresentado anteriormente, as ações são descritas utilizando-se uma notação de primeira ordem, o que é facilmente interpretado pelo ser humano mas dificilmente pelo computador. Então faz-se necessário um processo para transformar o conjunto de regras em primeira ordem para um conjunto de regras proposicionais que são facilmente interpretadas computacionalmente.

As informações de um problema de planejamento descritas em *PDDL* são normalmente armazenadas em arquivos, separados em arquivos de domínio e problema. Após obtê-las, utilizando-se de um analisador sintático, o planejador efetua a instanciação de todas as ações com todos os objetos gerando um conjunto de regras proposicionais.

Esse processo consiste em fazer a combinação de cada parâmetro de cada ação com todos os objetos possíveis. Como na definição dos parâmetros das ações foram definidos os tipos referentes, os objetos possíveis serão os que apresentam o mesmo tipo requerido. Por exemplo, dados os parâmetros da ação *carregar*: um pacote, um caminhão e uma localidade; e os objetos: *pacote*1 e *pacote*2 sendo do tipo *pacote*; *caminhão*1 do tipo *caminhão*; e *loja*1, *loja*2 e *depósito* como sendo do tipo *localidade*; faz-se a combinação dos objetos compatíveis com cada parâmetro, gerando:

carregar*(*pacote*1*,caminhão*1*,loja*1*)*=

    *(*{em*(*caminhão*1*,loja*1*)*∧em*(*pacote*1*,loja*1*)*},

    {dentro*(*pacote*1*,caminhão*1*)*},

    {em*(*pacote*1*,loja*1*)*}*)*,

carregar*(*pacote*1*,caminhão*1*,loja*2*)*=

    *(*{em*(*caminhão*1*,loja*2*)*∧em*(*pacote*1*,loja*2*)*},

    {dentro*(*pacote*1*,caminhão*1*)*},

    {em*(*pacote*1*,loja*2*)*}*)*,

carregar*(*pacote*1*,caminhão*1*,depósito*)*=

    *(*{em*(*caminhão*1*,depósito*)*∧em*(*pacote*1*,depósito*)*},

    {dentro*(*pacote*1*,caminhão*1*)*},

    {em*(*pacote*1*,depósito*)*}*)*,

carregar*(*pacote*2*,caminhão*1*,loja*1*)*=

*({em(caminhão1,loja1)∧em(pacote2,loja1)},*

       *{dentro(pacote2,caminhão1)},*

       *{em(pacote2,loja1)}),*

carregar*(pacote2,caminhão1,loja2)=*

       *({em(caminhão1,loja2)∧em(pacote2,loja2)},*

       *{dentro(pacote2,caminhão1)},*

       *{em(pacote2,loja2)}),*

carregar*(pacote2,caminhão1,depósito)=*

       *({em(caminhão1,depósito)∧em(pacote2,depósito)},*

       *{dentro(pacote2,caminhão1)},*

       *{em(pacote2,depósito)})*

Como foi citado anteriormente esse processo é realizado para cada ação existente no problema gerando ainda as ações instanciadas:

descarregar*(pacote1,caminhão1,loja1)=*

       *({em(caminhão1,loja1)∧dentro(pacote1,caminhão1)},*

       *{em(pacote1,loja1)}),*

       *{dentro(pacote1,caminhão1)},*

descarregar*(pacote1,caminhão1,loja2)=*

       *({em(caminhão1,loja2)∧dentro(pacote1,loja2)},*

       *{em(pacote1,loja2)}),*

       *{dentro(pacote1,caminhão1)},*

descarregar*(pacote1,caminhão1,depósito)=*

       *({em(caminhão1,depósito)∧dentro(pacote1,depósito)},*

       *{em(pacote1,depósito)}),*

       *{dentro(pacote1,caminhão1)},*

descarregar*(pacote2,caminhão1,loja1)=*

       *({em(caminhão1,loja1)∧dentro(pacote2,loja1)},*

       *{em(pacote2,loja1)}),*

       *{dentro(pacote2,caminhão1)},*

descarregar*(pacote2,caminhão1,loja2)=*

       *({em(caminhão1,loja2)∧dentro(pacote2,loja2)},*

       *{em(pacote2,loja2)}),*

       *{dentro(pacote2,caminhão1)},*

descarregar*(pacote2,caminhão1,depósito)=*

       *({em(caminhão1,depósito)∧dentro(pacote2,depósito)},*

       *{em(pacote2,depósito)}),*

       *{dentro(pacote2,caminhão1)},*

dirigir*(caminhão,loja1,loja2)=*

       *({em(caminhão,loja1)}, {em(caminhão,loja2)}, {em(caminhão,loja1)}),*

dirigir*(caminhão,loja1,depósito)=*

       *({em(caminhão,loja1)}, {em(caminhão,depósito)}, {em(caminhão,loja1)}),*

dirigir*(caminhão,loja2,loja1)=*

*({em(caminhão,loja2)}, {em(caminhão,loja1)}, {em(caminhão,loja2)}),*

dirigir*(caminhão,loja2,depósito)=*

*({em(caminhão,loja2)}, {em(caminhão,depósito)}, {em(caminhão,loja2)}),*

dirigir*(caminhão,depósito,loja1)=*

*({em(caminhão,depósito)}, {em(caminhão,loja1)}, {em(caminhão,depósito)}),*

dirigir*(caminhão,depósito,loja2)=*

*({em(caminhão,depósito)}, {em(caminhão,loja2)}, {em(caminhão,depósito)})*

É importante lembrar que se na definição das ações não estivesse definido o tipo dos parâmetros poderiam ocorrer instanciações que representariam ações incoerentes ou impossíveis do tipo dirigir um depósito de um pacote para um caminhão, ou ainda carregar um caminhão de um pacote para uma loja.

## 2.3  Estruturas de representação

O próximo passo para um planejador, depois de obtido o problema a partir de uma linguagem de descrição, por exemplo *PDDL*, é a construção de uma representação interna. Esta deve ser capaz de representar os estados e as ações que transformam um estado em outro.

O espaço de estados é a representação óbvia desse conceito pois é representado por um grafo onde os nós são os estados e as arestas são as transições entre os estados, ou seja, as ações. É construído a partir do estado inicial do problema e gerando novos estados a partir da execução das ações possíveis. Por exemplo, dado o estado inicial:

$$em(caminhão, loja2) \land em(pacote1, loja1)$$

tem-se que a ação *dirigir*$(caminhão, loja2, loja1)$ é válida, pois a proposição da lista $pre(o)$ - $em(caminhão, loja2)$ - está presente no estado apresentado. O resultado da aplicação dessa ação é ilustrado na figura 2.1:

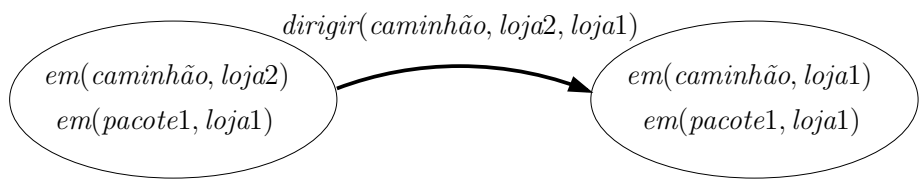

Figura 2.1: Aplicação da ação *dirigir*$(caminhão, loja2, loja1)$ do problema do comerciante segundo o espaço de estados.

Aplicando sucessivamente todas as ações permitidas torna-se teoricamente possível encontrar a seqüência de ações que transformam o estado inicial no estado objetivo, ou seja, um plano. A grande dificuldade encontrada é que na maioria dos problemas o espaço de busca gerado é muito grande, sendo inviável em termos de tempo de processamento ou espaço de armazenamento.

Apesar de outras representações terem sido propostas, neste trabalho analisaremos apenas o grafo de planos pois se trata da principal estrutura de representação dos dias de hoje [BF95]. As duas próximas seções mostram esta estrutura em mais detalhes. Por questões de apresentação, primeiramente veremos o grafo de planos relaxado.

## 2.3.1   O grafo de planos relaxado

O *grafo de planos relaxado* é formado por camadas, cada uma definida pelos nós pertencentes a ela. As camadas pares contêm os *nós proposições* que representam cada proposição possível para o problema a ser resolvido.

As camadas ímpares contêm os *nós ações*, que são as instâncias de ações cujas pré-condições são satisfeitas pelas proposições da camada anterior. Para facilitar a distinção entre proposições e ações, utilizaremos círculos para representar os nós proposições e retângulos para representar os nós ações.

As arestas do grafo conectam os nós proposições aos nós ações da camada posterior, ligando as pré-condições das ações às proposições correspondentes. Da mesma forma, as arestas ligam os efeitos de ações de uma camada às proposições correspondentes da camada seguinte.

O grafo apresenta ainda um caráter temporal representado pelas camadas de ações. Cada camada representa uma instância de tempo na aplicação das ações, assim, todas as ações presentes em uma determinada camada podem ser aplicadas simultaneamente.

Utilizaremos como exemplo o mesmo problema apresentado na seção anterior, do comerciante, mas, devido a limitação de espaço, será tratado apenas o problema envolvendo duas lojas, um caminhão e um pacote. O estado inicial é o *pacote* estar na *loja*1 e o *caminhão* na *loja*2. Observando a figura 2.2, que representa o grafo de planos relaxado do problema do comerciante representado apenas por 5 camadas, observamos que as quatro ações presentes na camada 1 são possíveis de execução no primeiro instante de tempo, e as oito ações da camada 3 são possíveis de execução no próximo instante de tempo.

Ao considerar o caráter temporal do grafo é necessário incluir uma ação de manutenção durante sua construção. Esta ação tem uma pré-condição e um efeito que é a própria pré-condição, ou seja, ela mantém a proposição $p$ válida também na próxima camada. Por exemplo a ação *manutenção(pacote,loja*1) faz com que o pacote ainda esteja na *loja*1 no próximo instante de tempo.

Para cada proposição existente na camada $i$ é aplicada a referida ação de manutenção ocasionando novamente a existência dessas proposições na camada $i + 2$. Esse processo garante que uma ação executada na camada $i$ seja novamente válida na próxima camada,

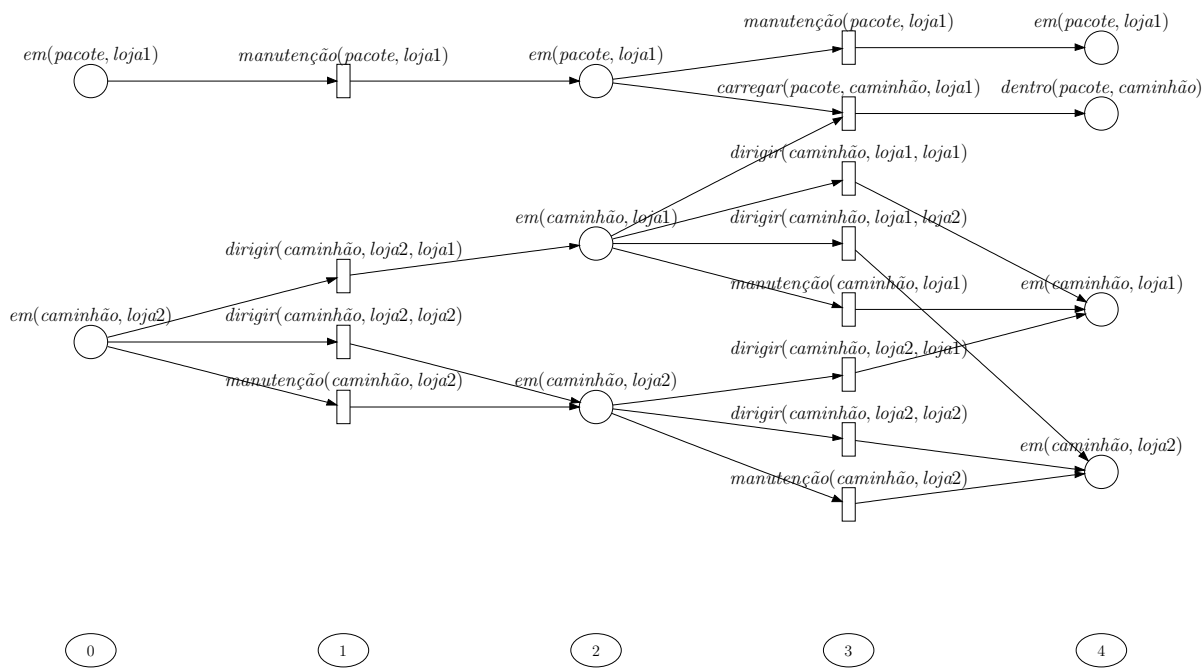

Figura 2.2: Grafo de planos relaxado para o problema do comerciante com 5 camadas.

ou seja, no próximo instante de tempo.

O grafo de planos relaxado é construído através de um processo de expansão que consiste em se adicionar ao grafo uma camada de nós com instâncias de ações que tenham suas pré-condições satisfeitas na camada anterior e inserir *somente os efeitos positivos* na camada posterior, o que é uma característica do problema de planejamento relaxado definido na seção 2.1.

O procedimento de construção do grafo é iniciado com a atribuição da primeira camada de proposições, que representa o estado inicial do problema. Na primeira expansão, as ações que possuem todas as pré-condições presentes no estado inicial são incluídas no grafo. Após a camada de ações, uma nova camada de proposições é incluída, esta contém as proposições de efeito positivos das ações que foram adicionadas na camada anterior.

A expansão ocorre até que uma camada de proposições contenha as proposições do estado final desejado ou até que uma camada de proposições obtida seja idêntica à camada de proposições anterior. A figura 2.3 mostra o grafo de planos relaxado para o problema do comerciante que tem como objetivo levar o *pacote* da *loja*1 para a *loja*2 sendo que o *caminhão* inicialmente está na *loja*2.

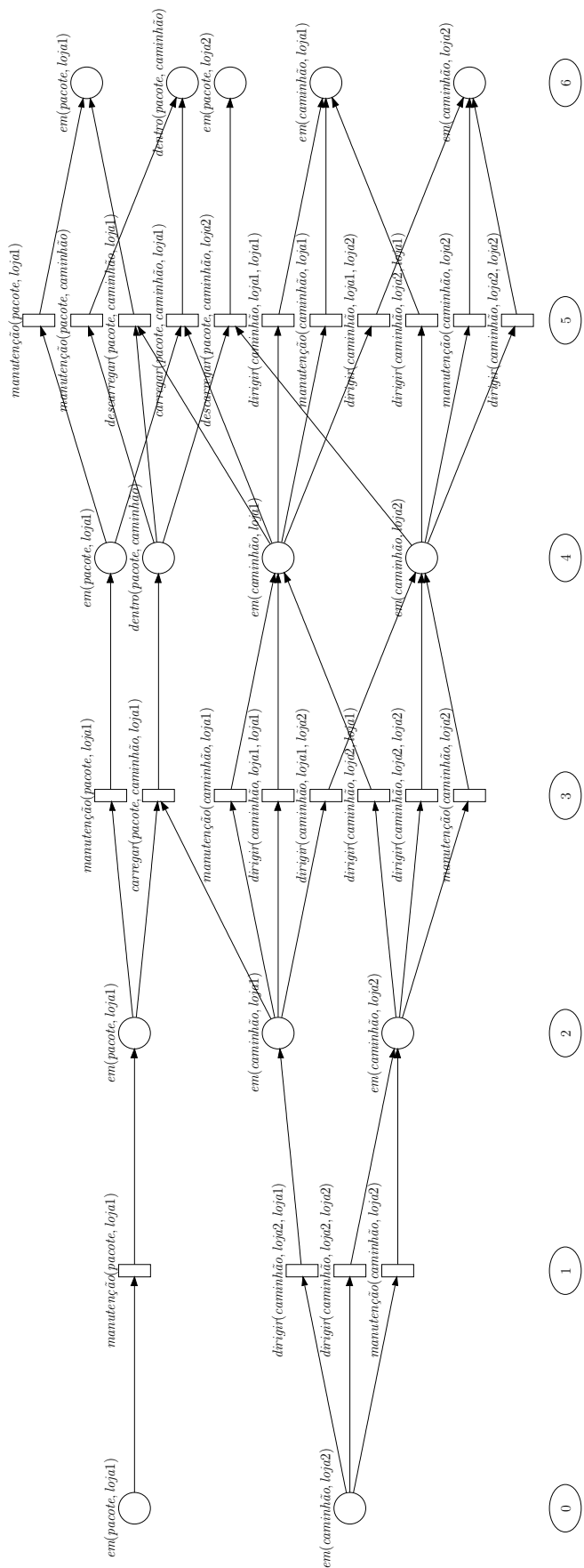

Figura 2.3: O grafo de planos relaxado para o problema do comerciante.

Pode-se observar que obtivemos um grafo com somente três camadas de ações o que sugere que o problema do comerciante poderia ser resolvido por três ações ou conjuntos de ações, se executadas em paralelo. Mas ao analisar o problema considerando as ações completamente, sem desprezar os efeitos de remoção, observa-se que essa expansão não é completa, pois ao dirigir o caminhão da *loja*2 para a *loja*1, não é removido a proposição do *caminhão* estar na *loja*2 possibilitando que, após carregado o *pacote*, basta descarregá-lo pois o *caminhão* também se encontraria na *loja*2.

Ao codificarmos o problema de planejamento considerando a lista de efeitos removidos encontraremos situações de inconsistências e conflitos, pois nem todas as ações numa camada são viáveis devido a contradições entre suas pré-condições e efeitos.

Por exemplo, a ação *dirigir*(*caminhão*, *loja*2, *loja*1) remove a proposição *em*(*caminhão*, *loja*2) que é pré-condição para *descarregar*(*pacote*, *caminhão*, *loja*2), caracterizando uma restrição entre estas ações: elas não podem ser executadas na mesma camada.

Na verdade existe uma série de outras restrições entre ações e proposições nas camadas do grafo de planos. Encontrar estas restrições não é simples. Blum e Furst definiram as chamadas relações de exclusão mútua (ou mutex), que são apresentadas a seguir.

## 2.3.2   O grafo de planos

O *grafo de planos* apresenta uma estrutura igual a do grafo de planos relaxado, diferenciando-se por não ser mais construído a partir de um problema de planejamento relaxado, mas sim considerando também os efeitos de remoção das ações. Quando considerado as ações com seus efeitos de remoção, encontraremos situações em que nem todas as ações numa camada são executáveis devido a contradições entre suas pré-condições e efeitos.

Retomando o exemplo do comerciante, temos que a ação *dirigir*(*caminhão*,*loja*1,*loja*2) remove a proposição *em*(*caminhão*,*loja*1) que é pré-condição para a execução da ação *carregar*(*pacote*,*caminhão*,*loja*1) indicando uma relação de exclusão mútua entre as duas ações. Esta característica reforça o caráter temporal do grafo, isto é, indica que estas duas ações não podem ser executadas ao mesmo tempo.

A relação de exclusão mútua entre ações (mutex) é representada por uma aresta que liga ações presentes em uma mesma camada. Esta relação é definida como segue:

**Definição:** Duas instâncias de ações *a* e *b* numa camada de ações *i* são mutuamente exclusivas se:

- um efeito de uma ação é a negação de um dos efeitos da outra (figura 2.4);

- um efeito de uma ação é a negação de uma pré-condição de outra (figura 2.5);

- as ações *a* e *b* têm pré-condições que são mutuamente exclusivas na camada *i* − 1 (figura 2.6). Este caso é chamado de *competição de necessidades*.

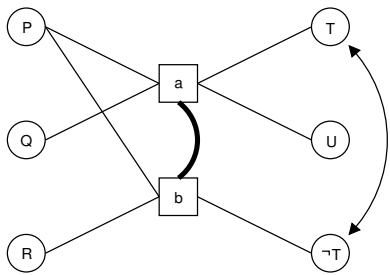

Figura 2.4: A ação $b$ tem como efeito a proposição $\neg T$ que é a negação de um efeito da ação $a$, portanto as ações $a$ e $b$ são mutuamente exclusivas na camada.

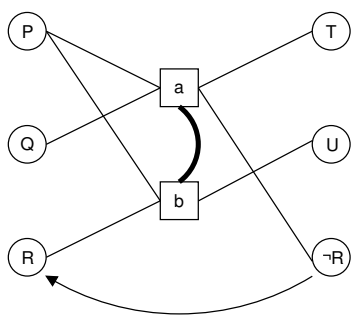

Figura 2.5: A ação $a$ tem como efeito a proposição $\neg R$ que é a negação de uma pré-condição da ação $b$, portanto as ações $a$ e $b$ são mutuamente exclusivas.

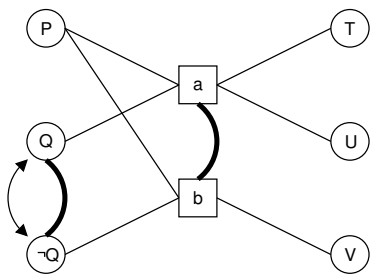

Figura 2.6: As ações $a$ e $b$ têm pré-condições que são mutuamente exclusivas na camada anterior $Q$ e $\neg Q$, portanto também são mutuamente exclusivas.

A relação de exclusão mútua também é definida para as proposições. Ela indica que as proposições assim relacionadas não podem ser obtidas simultaneamente na mesma camada, ou seja, somente uma delas pode ser utilizada. Essa relação é apresentada também como uma aresta que liga duas proposições em uma mesma camada, definida como segue:

**Definição:** Duas proposições $P$ e $Q$ numa camada de proposições $i$ são mutuamente exclusivas se todas as maneiras de obter as proposições são mutuamente exclusivas entre si, chamado *suporte inconsistente*. Ou seja, as ações da camada $i-1$ que têm como efeito estas proposições são duas a duas mutuamente exclusivas (figura 2.7).

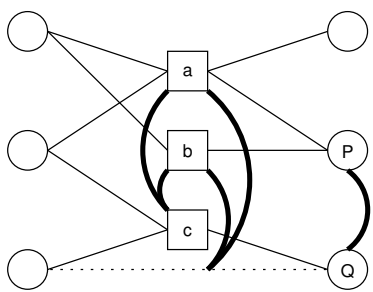

Figura 2.7: As proposições $P$ e $Q$ são mutuamente exclusivas devido às exclusões mútuas entre as ações que obtêm $P$ ($a$ e $b$) e as que obtêm $Q$ ($c$ e a ação de manutenção).

**Observação**: A definição de relação de exclusão mútua é recursiva, como no terceiro caso de exclusão entre ações: "as ações têm pré-condições que são mutuamente exclusivas na camada $i-1$". Portanto uma exclusão mútua numa determinada camada pode ter sido gerada por ações ou proposições de uma camada anterior. Um novo processo de expansão se faz necessário e é aqui que se nota a importância das ações de manutenção.

O grafo de planos construído considerando as relações de exclusão mútua descritas acima, apresentado na figura 2.8, possui uma representação mais completa comparado com o grafo de planos relaxado.

O processo de expansão é semelhante ao empregado no grafo de planos relaxado diferenciando-se apenas pela adição dos efeitos de remoção e pela finalização que ocorre quando o estado final está contido na última camada de proposições sem apresentar inconsistências e conflitos. Comparado com o grafo de planos relaxado, observa-se que foi necessário mais uma expansão do grafo para obter as proposições do estado final sem apresentar relação de exclusão mútua nas proposições do estado final.

Caso uma solução não seja encontrada, através de algum algoritmo de busca, uma nova expansão do grafo se faz necessária e um novo processo de busca é então executado no grafo expandido. Esse processo continua até que seja encontrada uma solução, processo apresentado em detalhes na seção 2.4.1.

Como o grafo de planos representa vários estados em uma mesma camada e, de forma similar, representa várias ações acontecendo em uma mesma camada, torna-se uma representação eficiente do espaço de busca. Esse foi um grande incentivo para o desenvolvimento de vários planejadores que utilizaram o grafo de planos como representação do espaço de busca. As seções seguintes apresentam os principais.

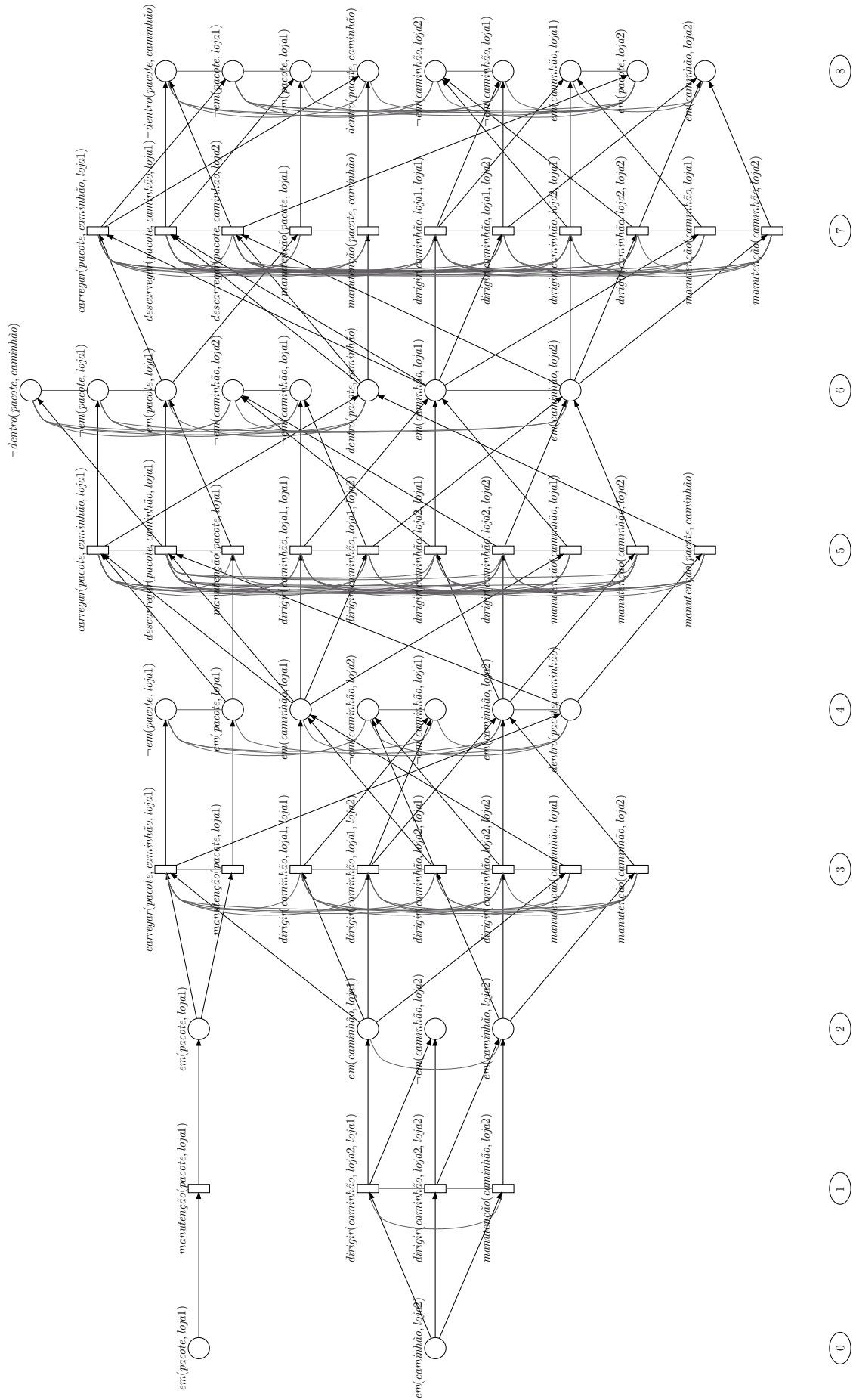

Figura 2.8: O grafo de planos para o problema do comerciante.

## 2.4 Planejadores baseados no grafo de planos

Nesta seção apresentaremos os principais planejadores que utilizam o grafo de planos para representar o espaço de busca. Os algoritmos variam segundo a técnica utilizada, compreendendo: busca exaustiva, satisfatibilidade, programação inteira, alcançabilidade e busca heurística. Eles são apresentados nas próximas seções.

### 2.4.1 Busca exaustiva no grafo de planos

O planejador *GRAPHPLAN*, apresentado por Blum e Furst [BF95], utiliza um algoritmo simples de busca exaustiva no grafo de planos gerado conforme apresentado na seção 2.3.2. Este grafo, contém inicialmente todas as proposições do estado final desejado, ou sub-metas, na sua última camada. Então, é executado o processo de extração da solução.

O objetivo da fase de extração da solução é obter um plano que seja solução para o problema. Esse processo utiliza-se de uma estratégia de *busca regressiva* que consiste em, a partir de cada sub-meta $m$ presente na camada $i$, encontrar uma ação $a$ na camada $i-1$ que tenha $m$ como efeito. Se $a$ é *consistente*, isto é, não apresenta conflitos com as outras ações escolhidas anteriormente na mesma camada, então passamos para a próxima sub-meta. Caso contrário voltamos e escolhemos outra ação na camada $i+1$ (*retrocesso*).

Se é encontrado um conjunto de ações consistente na camada $i-1$ que obtenham todas sub-metas, usamos o mesmo processo, recursivamente, para a camada $i-3$, considerando como sub-metas as pré-condições das ações escolhidas para a camada $i-1$, ou seja, as que formam a camada $i-2$. Senão é encontrado um conjunto de ações consistentes para a camada $i-3$, o *GRAPHPLAN* despreza as ações consideradas para a camada $i-1$ e escolhe outro conjunto. Caso não seja possível encontrar outro conjunto de ações consistentes, o processo falha e uma nova expansão do grafo é requerida. Isso indica que não é possível encontrar um caminho sem conflitos para alcançar os objetivos, ou seja, o número de camadas existentes nesse grafo é insuficiente sendo necessário mais uma expansão para tentar eliminar os conflitos existentes no caminho.

Caso as duas últimas camadas ímpares do grafo expandido sejam idênticas, ou seja, a última expansão não inseriu nenhuma proposição ou não eliminou nenhum conflito, não existe solução para o problema e nenhum plano solução poderá ser encontrado.

Quando a camada zero é alcançada utilizando conjuntos de ações consistentes nas camadas anteriores, temos uma solução para o problema de planejamento. Esse plano solução é formado pela união dos conjuntos de ações escolhidas, cuja ordem é apresentada pela camada, ou seja, as ações presentes na camada 1, são executadas antes das ações presentes na camada 3, e assim por diante.

Entre as ações que formam o plano solução, possivelmente encontra-se ações de manutenção. Estas ações devem ser desconsideradas pois somente permitem que as proposições estejam também presentes na camada seguinte, mas não representam nenhum significado real para a solução do problema.

Aplicando esse procedimento no grafo de planos gerado a partir do problema do comerciante (figura 2.8), temos que:

- primeiramente toma-se a única sub-meta *em(pacote,loja2)*;

- é encontrada a ação *descarregar(pacote,caminhão,loja2)* que tem a sub-meta como efeito e como é a primeira ação, não é mutex com nenhuma outra; não tem mais nenhuma sub-meta então passamos para a camada anterior considerando como sub-meta as proposições *dentro(pacote,caminhão)* e *em(caminhão,loja2)* que são pré-condições da ação considerada;

- escolhe-se então a ação *carregar(pacote,caminhão,loja2)* que gera a sub-meta *dentro(pacote,caminhão)*;

- escolhe-se então a ação *dirigir(caminhão,loja1,loja2)* que gera a sub-meta *em(caminhão,loja2)* mas esta não pode ser considerada pois é inconsistente com a primeira;

- escolhe outra ação, *dirigir(caminhão,loja2,loja2)*, mas esta também é inconsistente;

- escolhe ainda a ação *manutenção(caminhão,loja2)* mas também é inconsistente;

- assim volta a ação escolhida para a sub-meta anterior, *dentro(pacote,caminhão)*, e escolhe outra ação possível, a ação *manutenção(pacote,caminhão)*;

- então escolhe novamente a ação *dirigir(caminhão,loja1,loja2)* que agora é consistente; se considera agora as proposições *em(caminhão,loja1)* e *dentro(pacote,caminhão)* como as novas sub-metas;

- escolhe-se a ação *carregar(pacote,caminhão,loja1)* que gera a proposição *dentro(pacote,caminhão)*;

- escolhe-se a ação *manutenção(caminhão,loja1)* que gera a proposição *em(caminhão,loja1)* e é consistente; se considera então as proposições *em(pacote,loja1)* e *em(caminhão,loja1)* como as sub-metas;

- escolhe-se a ação *manutenção(pacote,loja1)* que gera a proposição *em(pacote,loja1)*;

- escolhe-se a ação *dirigir(caminhão,loja2,loja1)* que gera a proposição *em(caminhão,loja1)* e é consistente;

- chega-se a camada inicial (0) indicando que foi encontrado um plano solução para este problema.

O próximo processo é a extração do plano solução, que nada mais é que a ordenação das ações escolhidas de forma inversa à encontrada durante a busca, ou seja, da última ação escolhida até a primeira. As camadas indicam a ordem que estas ações devem ser executadas. Como já foi informado, ações em uma mesma camada podem ser executadas em paralelo.

Assim, o plano solução para o problema do comerciante é definido como:

- Camada 1: *dirigir(caminhão,loja2,loja1)*, *manutenção(pacote,loja1)*;

- Camada 3: *carregar(pacote,caminhão,loja1)*, *manutenção(caminhão,loja1)*;

- Camada 5: *dirigir(caminhão,loja2,loja1)*, *manutenção(pacote,caminhão)*;

- Camada 7: *descarregar(pacote,caminhão,loja2)*;

Como as ações de manutenção são necessárias somente para a representação, o próximo passo é eliminá-las, obtendo-se um plano solução:

$P_s$ =<dirigir*(caminhão,loja2,loja1)*,carregar*(pacote,caminhão,loja1)*,
    dirigir*(caminhão,loja2,loja1)*,descarregar*(pacote,caminhão,loja2)*>

A figura 2.9 mostra o grafo de planos para o problema do comerciante com os nós que fazem parte da solução marcados, tanto das ações que formam o plano quanto as proposições utilizadas.

Os autores do *GRAPHPLAN*, em [BF95], apresentam alguns resultados comparativos com mais três planejadores da época, dois planejadores de ordem total: *PRODIGY* [CBE+92] e *PRODIGY-SABA* [CBE+92] e um planejador de ordem parcial, *UCPOP* [PW92]. A figura 2.10, apresenta os resultados obtidos para problemas do domínio do foguete (*rocket*) que consiste em distribuir $n$ caixas em três localidades utilizando dois foguetes.

Devido ao fado do *GRAPHPLAN* ser implementado em C e os outros três em LISP, os testes foram realizados em máquinas diferentes para compensar a lentidão do LISP. O *PRODIGY* e o *UCPOP* foram executados em máquinas mais rápidas e com mais memória objetivando compensar a diferença de implementação.

O *GRAPHPLAN* apresentou excelentes resultados mesmo com a diferença entre as máquinas, embora essa diferença dificilmente supra as diferentes características da linguagem, estrutura e implementação dos planejadores citados.

Mesmo apresentando bons resultados, o *GRAPHPLAN* apresenta limitações em problemas maiores pois sua estrutura de implementação requer mais memória que o disponível para armazenar o problema, impossibilitando-o que resolva problemas complexos. O método exaustivo na fase de extração da solução gera problemas semelhantes aos encontrados nos primeiros planejadores também inviabilizando-o para vários problemas.

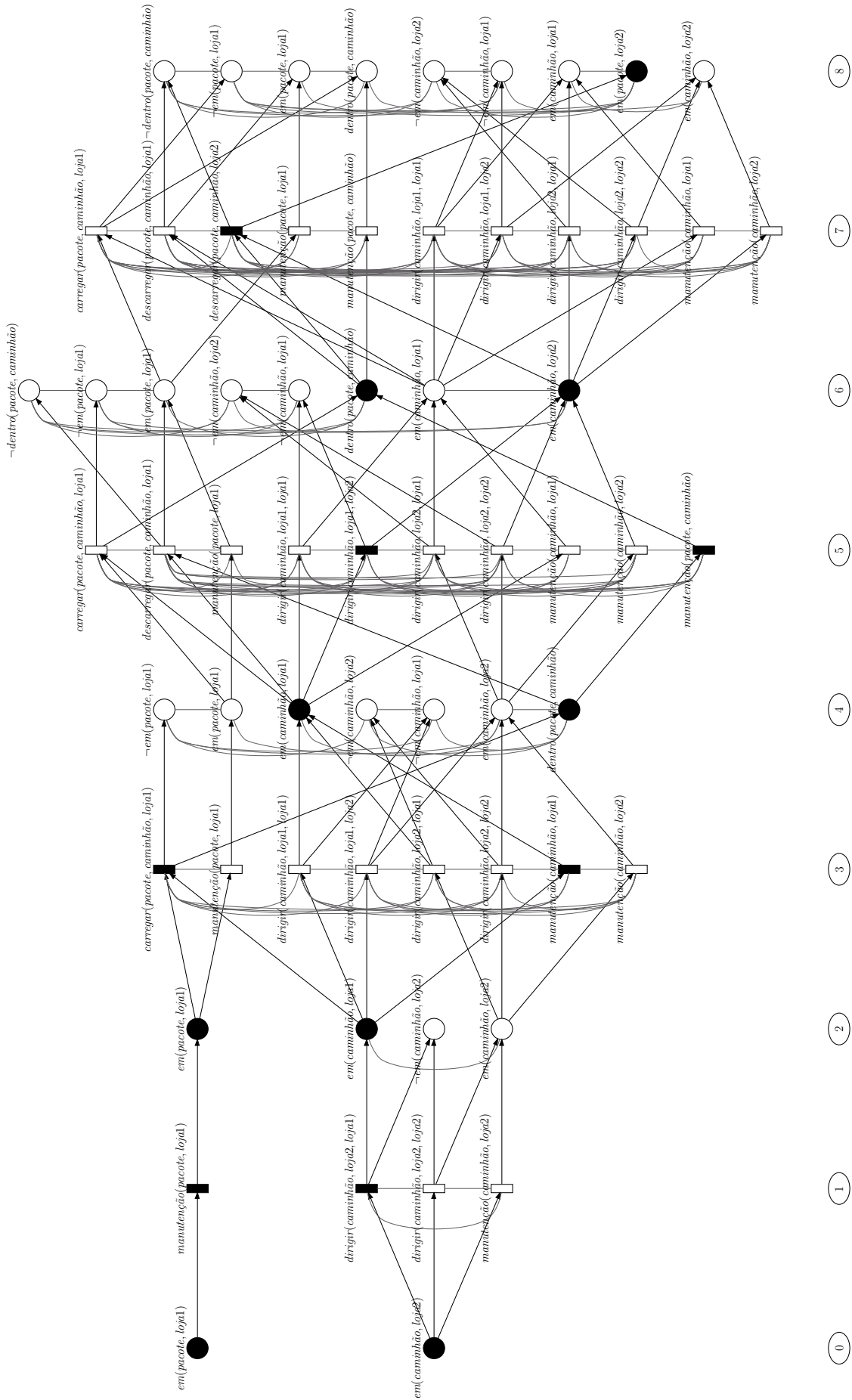

Figura 2.9: A solução para o problema do comerciante utilizando a busca exaustiva no grafo de planos.

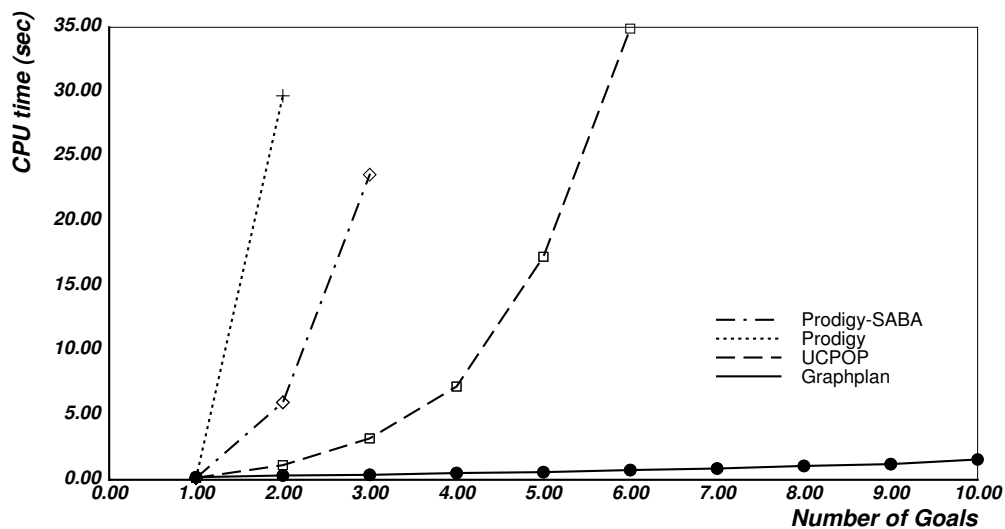

Figura 2.10: Resultados comparativos para resolução de problemas do domínio do foguete (*rocket*).

Tendo esse cenário que foi utilizado para testes do *GRAPHPLAN*, apresentando a diferença de hardware para compensar a diferença das características de implementação, surgem alguns questionamentos: como avaliar e quantificar a rapidez de uma linguagem? A diferença que foi dada é o suficiente? E a diferença da estrutura implementada para a representação do problema? Rapidez é a única característica que deve ser considerada quando se analisa planejadores?

Um fato importante e que aqui deve ser citado é a grande diferença apresentada pelos planejadores *IPP*[KNHD97] e *STAN*[LF99] que são as duas principais implementações do *GRAPHPLAN*. O *IPP* apresenta a remoção de ações e proposições irrelevantes e a agenda de gerenciamento dos objetivos enquanto o *STAN* implementa a construção do grafo utilizando operações lógicas binárias em vetores binários. Eles são um exemplo de como características de implementação influenciam no desempenho do planejador e por isso não podem ser desconsideradas.

## 2.4.2 Grafo de planos como pré-processamento

Graças à otimização da estrutura utilizada para representação do espaço de busca, a construção do grafo de planos pode ser vista como uma fase de pré-processamento para procedimentos de busca alternativos à busca exaustiva empregada no *GRAPHPLAN* original. Nesta seção mostraremos duas técnicas: tradução para satisfatibilidade e tradução para redes de Petri.

## Satisfatibilidade a partir do grafo de planos

O planejador *BLACKBOX* foi proposto por Kautz e Selman [KS99] em 1998. Os autores tiveram a idéia de unir a eficiência dos algoritmos para satisfatibilidade, apresentados no *SATPLAN* [KS92] com o grafo de planos (seção 2.3.2) que, como vimos, otimiza a representação do espaço de busca.

O *SATPLAN* resume-se em um processo que, a partir de um problema de planejamento descrito em *PDDL* (seção 2.2), gera um problema de satisfatibilidade (SAT) que é então resolvido através de algum método conhecido. Esses métodos buscam por uma atribuição de valores que resulte na satisfatibilidade da fórmula em Forma Normal Conjuntiva, por exemplo: Walksat [SKC94], SatZ, SatZ-Rand [LA97] ou Rel_sat [BS97]. O *BLACKBOX* se diferencia do *SATPLAN* pois obtém a instância SAT a partir do grafo de planos.

Uma fórmula ou instância SAT, é um conjunto de restrições, ou literais, que assumem o valor verdadeiro ou falso. Ao tratarmos planejamento como satisfatibilidade, um plano corresponde a um assinalamento verdade para um conjunto de restrições que representam o estado inicial, o estado final e os axiomas do domínio.

Como um problema de planejamento tem uma componente temporal inerente, pois um plano é uma seqüência cronológica de ações, e o problema de satisfatibilidade é um problema estático de valoração, o tempo consiste de um número fixo e discreto de instâncias.

Uma proposição corresponde a um literal variante do tempo pertencente a um instante particular (ex.: $limpo(m\tilde{a}os, 0)$, indica que as mãos estão limpas no instante 0), ou a uma ação que ocorre no instante especificado e finaliza no instante seguinte (ex.: $cozinhar(jantar, 1)$, indica que ocorre a ação *cozinhar* no instante 1).

Restrições sobre proposições e ações são escritas como *esquemas*, que são instanciados pelos objetos e pelas instâncias de tempo definidos em um problema particular. O tamanho máximo de um plano é então fixado no tempo máximo instanciado; se esta quantidade não é conhecida, é executada uma busca binária nas instanciações de vários tamanhos, até encontrar a menor instância em que a solução é encontrada.

A instância encontrada é resolvida por algum método SAT conhecido, que geralmente apresenta-se mais eficiente comparado com métodos de prova de teorema. Mas como as instâncias geradas a partir do espaço de estados eram grandes, utilizou-se o grafo de planos para reduzir o tamanho das instâncias. Outra otimização foi a utilização do número de camadas do grafo como tamanho inicial para as instâncias, valor este necessário para a construção das instâncias.

O algoritmo do *BLACKBOX* é definido por três fases:

1. Um problema de planejamento é representado como um grafo de planos;

2. O grafo de planos é convertido em uma instância SAT;

3. A instância SAT é resolvida por qualquer um dos métodos para SAT;

Uma vez já apresentado o grafo de planos (seção 2.3.2) descreveremos o método de conversão em uma instância SAT proposto em 1996 por Kautz e Selman [KS96]. Essa conversão inicia na última camada do grafo e termina na primeira camada. Usaremos o problema do foguete (*rocket*) apresentado por Blum e Furst [BF95] como um exemplo.

Esse problema consiste em utilizar um robô para levar peças de um lugar para outro. Possui três ações: *carregar*, que carrega o robô com alguma peça; *descarregar*, que descarrega a peça; e *mover*, que desloca o robô de uma localização para outra. Existem: um robô $R$, duas peças $A$ e $B$, uma localização inicial $L$ e um destino $P$. Assim "*carregar*$(A, R, L, i)$" significa "carregue $A$ em $R$ na localização $L$ no tempo $i$", e "*mover*$(R, L, P, i)$" significa "mova $R$ de $L$ para $P$ no tempo $i$".

O processo de conversão do grafo de planos em SAT é descrito como segue:

1. O estado inicial pertence a camada 1 e o estado final pertence a camada com maior valor;

2. Cada proposição $P$ da camada de proposição $i$ implica na disjunção de todos as ações na camada $i - 1$ que tem $P$ como um efeito de adição, por exemplo:

$$dentro(A, R, 3) \rightarrow (carregar(A, R, L, 2) \vee$$
$$carregar(A, R, P, 2) \vee$$
$$manutenção(dentro(A, R), 2))$$

3. Ações implicam em suas pré-condições, por exemplo:

$$carregar(A, R, L, 2) \rightarrow (em(A, L, 1) \wedge em(R, L, 1));$$

4. Ações conflitantes (mutex) são mutuamente exclusivas, por exemplo:

$$\neg carregar(A, R, L, 2) \vee \neg mover(R, L, P, 2).$$

A partir da instância gerada pela compilação para SAT, um processo de simplificação é aplicado à fórmula final obtida, com o objetivo de reduzi-la.

Aplicando a regra dos "axiomas de persistência" [Haa87] nos axiomas gerados a partir do (2) com os axiomas gerados a partir do (3) que apresentam ações de manutenção possibilita uma redução significativa nas instâncias. Por exemplo, resolvendo o axioma:

$$dentro(A, R, 3) \rightarrow$$
$$(carregar(A, R, L, 2) \vee carregar(A, R, P, 2) \vee manutenção(dentro(A, R), 2))$$

com o axioma:

$$manutenção(dentro(A, R), 2) \rightarrow dentro(A, R, 1)$$

formamos o axioma de persistência:

$$(\neg dentro(A, R, 1) \wedge dentro(A, R, 3)) \rightarrow (carregar(A, R, L, 2) \vee carregar(A, R, P, 2)).$$

Segundo a teoria proposicional, qualquer variável pode ser eliminada através da execução de todas resoluções possíveis sobre aquela variável e removendo todas as cláusulas contendo essa variável. Assim, qualquer subconjunto de literais, se representam ações ou proposições, podem em princípio ser eliminadas.

Em geral o processo de resolução obtém um aumento exponencial no tamanho da instância, mas para a codificação baseada no grafo de planos a eliminação das proposições ocasiona somente um aumento polinomial neste tamanho. Por isso é executado o processamento de eliminação das proposições, permanecendo apenas as ações.

O processo ocorre da seguinte forma:

- Dadas as proposições:

  $p_1$ e $p_2$

- Os axiomas das ações:

  $a_3 \rightarrow p_1$ ; $a_3 \rightarrow p_2$ ; $a_4 \rightarrow p_1$ ; $a_4 \rightarrow p_2$

- E os axiomas das proposições:

  $p_1 \rightarrow (a_1 \vee a_2)$ e $p_2 \rightarrow (a_1 \vee a_2)$

- Obtém-se:

  $a_3 \rightarrow (a_1 \vee a_2)$ e $a_4 \rightarrow (a_1 \vee a_2)$

Após a redução, a fórmula é então passada como entrada para um algoritmo que resolve o problema de satisfatibilidade. O algoritmo resolvedor busca, por uma atribuição de valores aos literais, a satisfatibilidade da fórmula. Se a fórmula é satisfável, esta valoração é traduzida usando a tabela de símbolos do compilador para um plano solução.

Kautz e Selman apresentam alguns resultados comparativos entre o *BLACKBOX*, *SATPLAN*, *GRAPHPLAN* e *IPP* [KNHD97] (figura 2.11).

Podemos observar claramente que o *BLACKBOX* utilizando o SatZ-Rand apresentou os melhores resultados em todos os problemas apresentados, uma vez que o *SATPLAN* deve ser considerado em todos os valores o processo de criação das instâncias. Mas o *BLACKBOX* apresenta o mesmo problema que o *GRAPHPLAN*, o crescimento exponencial do grafo de planos limitando sua execução para alguns problemas. Outra característica interessante de se observar é que a comparação é feita apenas nos domínios do foguete (*rocket*) e logística e que não comparam com outros planejadores tais como *UCPOP*.

| problem | parallel time | Blackbox | | | Graphplan | IPP | SATPLAN | | | |
|---|---|---|---|---|---|---|---|---|---|---|
| | | walksat | satz | satz-rand | | | create | walksat | satz | satz-rand |
| rocket.a | 7 | 3.2 sec | 5 sec | 5 sec | 3.4 min | 28 sec | 42 sec | 0.02 sec | 0.3 sec | 2 sec |
| rocket.b | 7 | 2.5 sec | 10 sec | 5 sec | 8.8 min | 55 sec | 41 sec | 0.04 sec | 0.3 sec | 1 sec |
| log.a | 11 | 7.4 sec | 5 sec | 5 sec | 31.5 min | 1 hour | 1.2 min | 2 sec | 1.7 min | 4 sec |
| log.b | 13 | 1.7 min | 7 sec | 7 sec | 12.7 min | 2.5 hour | 1.3 min | 3 sec | 0.6 sec | 7 sec |
| log.c | 13 | 14.9 min | 9 sec | 9 sec | — | — | 1.7 min | 2 sec | 4 sec | 0.8 sec |
| log.d | 14 | — | 52 sec | 28 sec | — | — | 3.5 min | 7 sec | 1.8 hour | 1.6 min |

Figura 2.11: Resultados comparativos para resolução de problemas do domínio do foguete (*rocket*) e logística [KNHD97].

## Alcançabilidade a partir do grafo de planos

O planejador *PETRIPLAN* [SCK00] segue uma estrutura semelhante à apresentada pelo *BLACKBOX*, ainda utilizando o grafo de planos como representação do espaço de busca, mas substituindo a codificação deste para um problema SAT por uma codificação do problema de planejamento como um problema de alcançabilidade de sub-marcação em uma rede de Petri [Mur89]. Redes de Petri são comumente utilizadas para representar paralelismo, concorrência, conflito e relações causais em sistemas dinâmicos de eventos discretos. A idéia dos autores é poder integrar estas noções no planejador.

O *PETRIPLAN* é um algoritmo definido por três fases:

1. Um problema de planejamento é codificado em um grafo de planos;

2. O grafo de planos é convertido para uma rede de Petri;

3. O problema de alcançabilidade em uma rede de Petri é resolvido utilizando sistemas específicos para programação inteira.

Se nenhuma solução for encontrada para o problema de programação inteira na terceira fase, o algoritmo volta para a primeira fase e uma nova expansão do grafo é realizada. Este laço continua até que uma solução seja encontrada na fase 3 ou até que o grafo de planos não possa ser mais expandido. Quando uma solução é encontrada na fase 3 ela é convertida para um plano que resolve o problema original de planejamento.

Como vimos anteriormente, o grafo de planos é um grafo ordenado formado por camadas alternadas de proposições e ações, ambas indexadas por valor único que pode ser interpretado como uma instância de tempo. Os arcos partem de cada proposição até as ações da próxima camada que a determinada proposição como uma pré-condição, e similarmente parte de cada ação até seus efeitos presentes na próxima camada. Tem-se ainda as ações de manutenção que simplesmente tem uma proposição tanto como pré-condição como efeito.

Uma rede de Petri é também representada por um grafo, onde círculos representam lugares, barras verticais representam transições e os arcos orientados representam funções de incidência de entrada e saída de uma transição. Pequenos círculos pretos no interior dos lugares, chamados de marcas, representam a marcação da rede. Os lugares descrevem estados e as transições são responsáveis pelas mudanças de estado, elas removem e inserem marcas nos lugares.

Assim a figura 2.12(a) representa três lugares ($a$, $b$ e $c$), duas transições ($x$ e $y$), cinco arcos, uma marca no lugar $a$ e duas no lugar $b$. Ao dispararmos a transição $x$ da rede representada pela figura 2.12(a) obtemos a marcação resultante na figura 2.12(b).

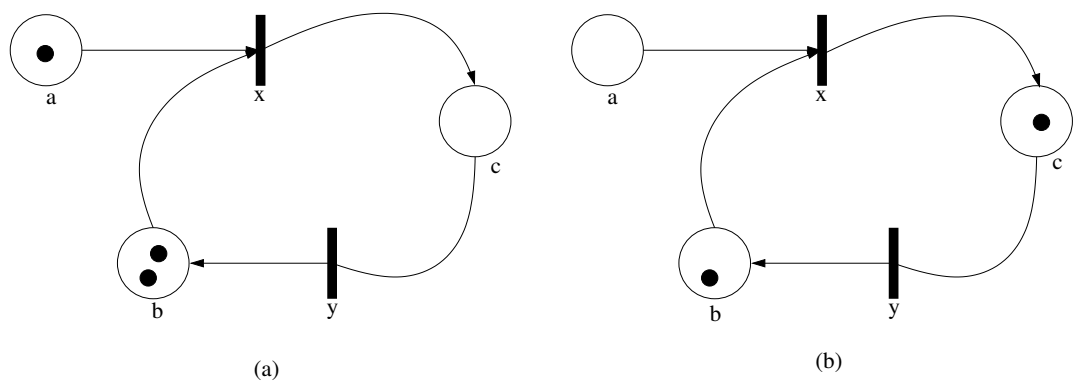

Figura 2.12: Representação gráfica de uma rede de Petri.

O *problema de alcançabilidade* para redes de Petri é definido como sendo o de encontrar uma seqüência de transições disparáveis para alcançar uma marcação objetivo a partir de uma marcação inicial. O *problema de alcançabilidade de sub-marcação* consiste em encontrar uma seqüência de transições disparáveis para alcançar um subconjunto de lugares marcados presentes na marcação objetivo.

A codificação de um grafo de planos em uma rede de Petri é um processo direto de conversão das estruturas do grafo de planos em estruturas equivalentes na rede de Petri. A seguir apresentamos estas traduções:

**Nós ação:** um nó ação do grafo é traduzido em uma única transição na rede de Petri, a figura 2.13 mostra à esquerda o nó do grafo e à direita a transição correspondente.

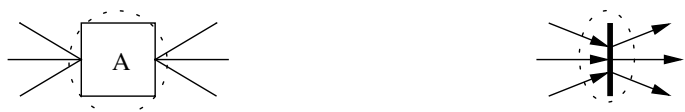

Figura 2.13: Tradução dos nós ação.

**Nós proposição:** um nó proposição é traduzido em um lugar e uma transição, com um arco do lugar para a transição. A figura 2.14 mostra esta tradução.

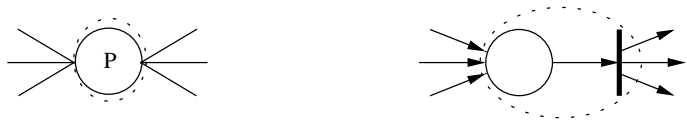

Figura 2.14: Tradução dos nós proposição.

**Arestas efeito:** uma aresta efeito é traduzida em um arco que vai da transição que representa o nó ação para o lugar representando o nó proposição. A figura 2.15 mostra esta tradução.

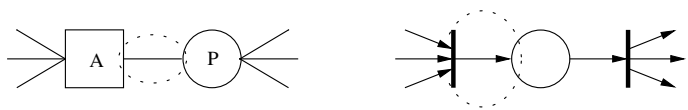

Figura 2.15: Tradução das arestas efeito.

**Arestas pré-condição:** uma aresta pré-condição é traduzida em um lugar com dois arcos: um vindo da transição que representa o nó proposição e outro que vai para a transição representando o nó ação. A figura 2.16 mostra esta tradução.

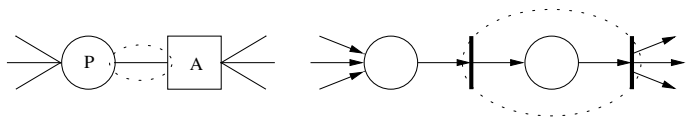

Figura 2.16: Tradução das arestas pré-condição.

**Exclusão mútua:** a relação binária de exclusão mútua é traduzida em um lugar com dois arcos saindo, um para cada transição que representa cada nó ação da relação, e uma marca neste lugar. A figura 2.17 mostra esta tradução.

As relações de exclusão mútua entre proposições não são representadas na rede, pois, estas são utilizadas somente durante a construção do grafo de planos para determinar as exclusões mútuas entre as ações do grafo.

O estado inicial do problema de planejamento é representado por marcas adicionadas em cada lugar que representa a camada zero do grafo de planos. As marcas nos lugares do estado inicial e as marcas nos lugares que controlam as relações de exclusão mútua definem a marcação inicial da rede.

Devido a amplitude do grafo de planos gerado para o problema do comerciante, utilizaremos aqui o problema do jantar [Wel99] para detalhar o funcionamento do algoritmo. Esse problema consiste de se preparar um jantar e embrulhar um presente, mas em silêncio, para não chamar atenção de quem está no segundo andar da casa. Tem-se a sua disposição as ações "cozinhar", que requer as mãos limpas e obtém o jantar mas remove o silêncio, e "embrulhar", que requer silêncio e obtém o presente embrulhado. O estado inicial é ter as mãos limpas e estar em silêncio. A figura 2.18 mostra o grafo de planos para este problema.

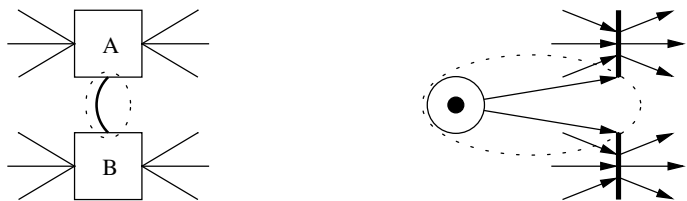

Figura 2.17: Tradução das relações de exclusão mútua.

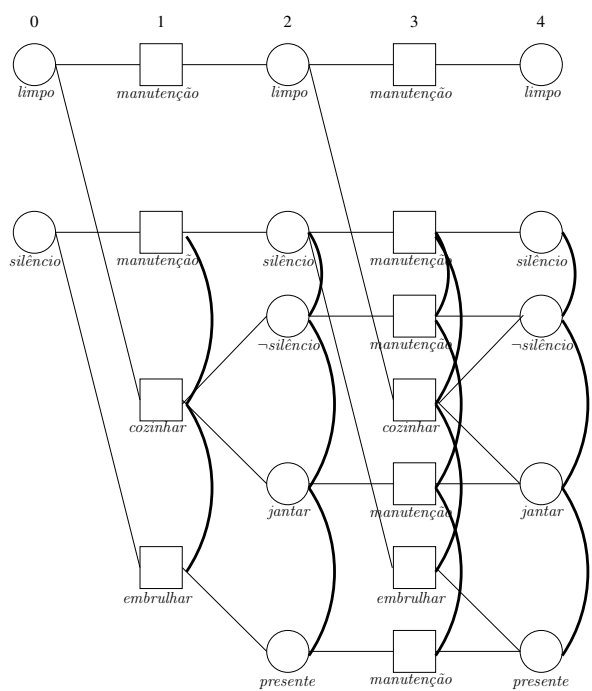

Figura 2.18: O grafo de planos para o problema do jantar.

A figura 2.19 mostra a rede de Petri obtida pela tradução do problema do jantar apresentado na figura 2.18.

O estado final do problema de planejamento é representado pela sub-marcação da rede que contém marcas nos lugares que representam os nós proposições do estado final. A figura 2.20 mostra uma rede de Petri com uma marcação que contém o estado final para o problema do jantar.

Para resolver o problema de alcançabilidade de sub-marcação, como se conhece apenas o subconjunto de lugares que representam as proposições do estado final desejado, deve-se encontrar a seqüência de transições que sai dos lugares que representam o estado inicial e chegam aos lugares que representam o estado objetivo. Considerando apenas as transições que representam ações nesta seqüência, temos um plano solução para problema original.

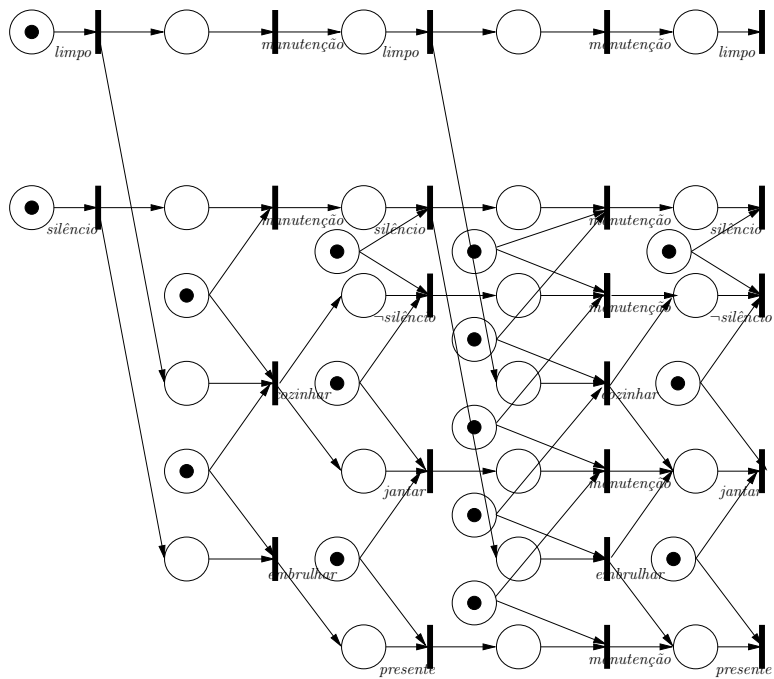

Figura 2.19: A rede de Petri com a marcação inicial para o problema do jantar.

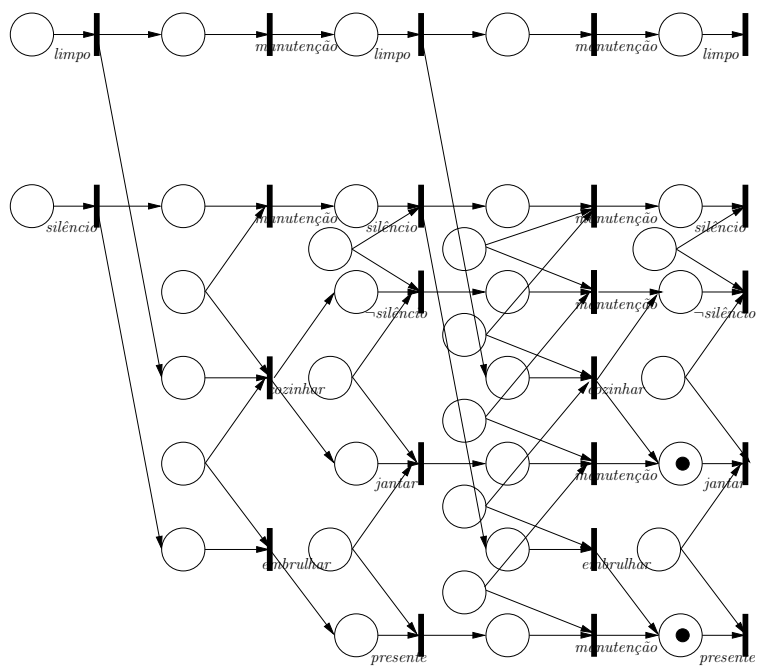

Figura 2.20: A rede de Petri com a marcação final desejada para o problema do jantar.

Os autores do *PETRIPLAN* apresentam alguns resultados comparativos entre o *PE-TRIPLAN*, *BLACKBOX*, *HSP* [BG98], *IPP* [KNHD97] e *STAN* [LF99]. Estes resultados são apresentados nas tabelas 2.1 e 2.2 e mostram o número de problemas resolvidos por cada planejador e o tempo médio gasto por problema.

Tabela 2.1: Número de problemas resolvidos por planejador.

| Domínio | *PETRIPLAN* | *BLACKBOX* | *HSP* | *IPP* | *STAN* | Problemas |
|---------|-------------|------------|-------|-------|--------|-----------|
| Gripper-1 | 2 | 2 | 11 | 4 | 4 | 20 |
| Logistics-1 | 1 | 3 | 4 | 5 | 2 | 30 |
| Mystery-1 | 4 | 13 | 16 | 13 | 14 | 30 |
| Mprime-1 | 0 | 15 | 23 | 16 | 8 | 30 |
| Logistics-2 | 2 | 3 | 5 | 4 | 3 | 5 |
| Mprime-2 | 0 | 4 | 4 | 4 | 4 | 5 |
| Grid-2 | 0 | 1 | 0 | 3 | 1 | 5 |
| Total | 9 | 41 | 63 | 49 | 36 | 125 |

Tabela 2.2: Tempo médio por problema resolvido por planejador. Os valores são apresentados em milisegundos e representam o tempo total de execução de um planejador para um determinado problema. Os campos marcados com "x" indicam que o planejador não conseguiu encontrar uma solução para o problema dentro dos limites de tempo e memória estabelecidos.

| Domínio | *PETRIPLAN* | *BLACKBOX* | *HSP* | *IPP* | *STAN* |
|---------|-------------|------------|-------|-------|--------|
| Gripper-1 | 1353707,00 | 4140,50 | 2126,36 | 37675,00 | 789313,25 |
| Logistics-1 | 4189902,00 | 4573,67 | 12171,00 | 716734,00 | 25272,50 |
| Mystery-1 | 3835687,00 | 2084,61 | 12078,19 | 8077,69 | 762,93 |
| Mprime-1 | x | 2538,40 | 118732,87 | 22396,87 | 1398,50 |
| Logistics-2 | 3480784,00 | 529,67 | 10036,60 | 17438,25 | 9037,33 |
| Mprime-2 | x | 1263,25 | 1767,50 | 6532,25 | 5205,50 |
| Grid-2 | x | 11035,00 | x | 32064,33 | 3666,00 |
| Total | 2392250,00 | 2557,44 | 48467,21 | 66964,63 | 121939,03 |

Podemos observar que o *PETRIPLAN* apresentou um tempo maior para os problemas que resolveu, e ainda não conseguiu resolver vários. Mas esse fato foi justificado devido a implementação deste ser utilizando uma biblioteca de resolução de PI não eficiente. Apesar do baixo desempenho, os resultados obtidos mostram que o algoritmo resolve corretamente os problemas, mas limitado principalmente por problemas na implementação.

O *PETRIPLAN* apresenta algumas deficiências sobre implementação que podem ser exploradas como a substituição da biblioteca de PI, explorar variações na tradução para redes de Petri e usar outras técnicas para resolver o problema de alcançabilidade.

Observamos novamente a dificuldade em se comparar os planejadores pois são analisados somente os resultados finais não apresentando as características de cada processo. Assim, pode-se questionar algumas informações importantes para a análise, como: os problemas que não foram resolvidos pelo *PETRIPLAN* apresentaram quais deficiências? Quais as particularidades do planejador que o inviabilizaram para esses problemas?

### 2.4.3 Busca heurística com base no grafo de planos

O planejador Fast-Forward (*FF*), foi desenvolvido por Jörg Hoffmann e Bernhard Nebel [HN01] e é baseado em dois planejadores: o *GRAPHPLAN* [BF95] e o *HSP* [BG98]. Do *GRAPHPLAN* foi utilizado sua representação, o grafo de planos (seção 2.3.2) e do *HSP* foi utilizado seu procedimento de busca heurística.

O *FF* realiza uma busca progressiva no espaço de estados, utilizando uma função heurística que estima as distâncias para o estado final pelo comprimento de uma solução aproximada para um problema de planejamento relaxada. O *FF* diferencia-se do *HSP* por possuir uma nova estratégia de busca local baseada em uma função heurística mais sofisticada, e duas otimizações.

O método de busca empregado pelo *FF* é o *subida de encosta reforçado*, que combina busca local com sistemática pois utiliza o grafo de planos relaxado, seção 2.3.1, como função heurística para guiar a busca local.

As técnicas de otimizações compreendem: a identificação de um conjunto de ações sucessoras mais promissoras para cada nó de busca, e outra que elimina ramificações onde os objetivos são atingidos muito rapidamente.

O sistema *FF* funciona da seguinte maneira:

- Aplica-se o algoritmo *subida de encosta reforçado* até que o objetivo seja alcançado ou que este algoritmo falhe;

- Se o algoritmo falhou, despreze tudo o que foi feito e resolva o problema utilizando um algoritmo de busca heurística completo (*Best-first*).

A descrição do método *subida de encosta reforçado* é definida como: partindo de um estado $S$, avalie todos os seus sucessores diretos $S'$. Se nenhum dos sucessores tem uma heurística melhor que $S$, procure pelos sucessores dos sucessores e assim sucessivamente até encontrar um estado $S'$ com uma heurística melhor do que $S$. Então o caminho até $S'$ é adicionado ao plano corrente, e a busca continua com $S'$ como o novo estado inicial $S$ até encontrar o estado final, $h(S) = 0$. Se nenhum estado $S'$ foi encontrado, então o método *falha*.

A função heurística $h(S)$ utilizada pelo *FF* é obtida através do grafo de planos relaxado, onde todas as ações não possuem a lista de efeitos de remoção (apresentado em mais detalhes na seção 2.3.2). O plano extraído do grafo de planos relaxado com $m$ camadas é formado pelo conjunto de ações $O_i$ selecionadas na camada $i$, portanto o plano resultante é $\langle O_0, ..., O_{m-1} \rangle$. A função heurística do *FF* é estimada a partir do comprimento da linearização desse plano:

$$h(S) := \sum_{i=0}^{m-1} |O_i|$$

Como vimos anteriormente o *FF* utiliza algumas técnicas de otimização, obtidas através do plano relaxado que o *FF* extrai a partir da busca realizada para cada estado. Com a busca pode-se identificar os sucessores mais promissores e detectar a informação de ordenação dos objetivos.

A primeira técnica de otimização, a identificação de um conjunto de ações sucessoras mais promissoras $H(S)$ para um estado $S$, é definida pelo conjunto das ações que adicionam pelo menos um objetivo na primeira expansão do grafo de planos relaxado $G_1$:

$$H(S) := \{o \mid pre(o) \subseteq S, add(o) \cap G_1 \neq 0\}$$

Assim, a escolha da uma ação é sempre aquela que tiver o menor número de précondições.

O método para eliminar ramificações que alcancem objetivos antecipadamente, a segunda técnica de otimização, consiste de: se um plano relaxado $P$, gerado a partir de $S$, contém uma ação $o$, $o \in P$, que remove o objetivo $G$ ($G \in del(o)$ na versão não relaxada de $o$), então remove-se $S$ do espaço de busca, ou seja, eliminamos a ramificação partindo de $S$.

A figura 2.21 apresenta os resultados obtidos por vários planejadores na competição de 2000 [Bac00] resolvendo problemas do domínio de logística. Este é um problema clássico que envolve o transporte de pacotes entre cidades utilizando caminhões e aviões.

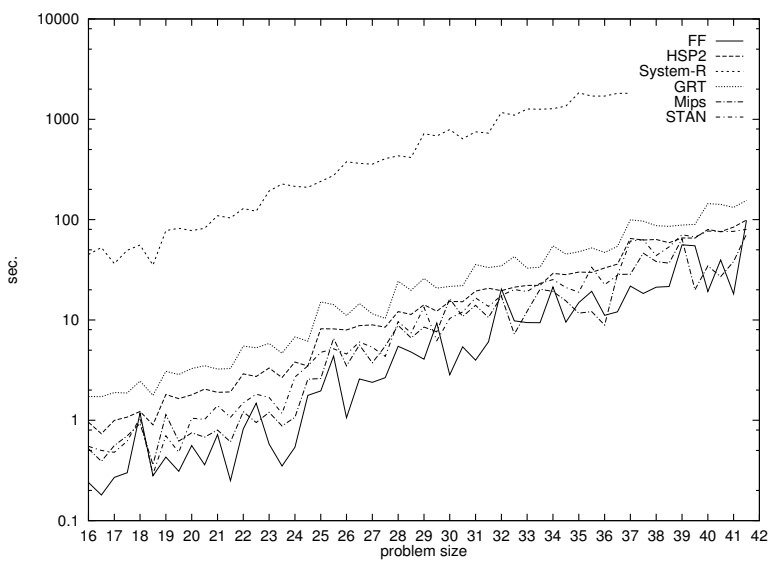

Figura 2.21: Resultados obtidos para o domínio de logística [Bac00].

A figura 2.22 apresenta os resultados obtidos resolvendo problemas do domínio mundo de blocos. Este também é um problema clássico cujo planejador precisa reorganizar um conjunto de blocos em uma posição específica usando braços de um robô.

Nestes dois grupos de testes apresentados podemos fazer algumas observações impor-tantes. Mesmo utilizando uma variação da estratégia de busca local *subida de encosta*, o *FF* ainda apresenta alguns problemas em domínios semelhantes ao mundo de blocos, uma vez que neste domínio os algoritmos de busca locais caem em mínimos locais. Em outros domínios, como é o caso do logística, o *FF* apresenta os melhores resultados, tanto relacionado ao tempo gasto quanto ao número de problemas solucionados.

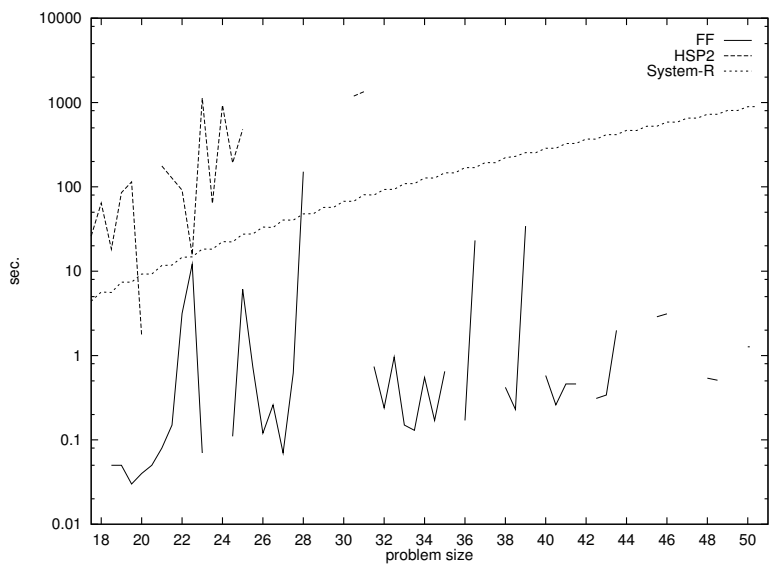

Figura 2.22: Resultados obtidos para o domínio mundo de blocos [Bac00].

Novamente observamos a dificuldade em se comparar os planejadores pois não são mais apresentados resultados dos domínios apresentados por outros planejadores, somente são apresentados os domínios nos quais este planejador apresenta resultados satisfatórios. Além de desconsiderar as diferenças de implementações.

## 2.5 As competições de planejadores

Devido ao grande número de planejadores existentes e a diversidade de implementações e metodologias, foi criada a Competição de Planejadores como um meio de se comparar os planejadores. A competição é um evento bienal que acontece na AIPS - *International Conference on Artificial Intelligence Planning Systems* e teve início em 1998. Tem o objetivo de promover o desenvolvimento de sistemas de planejamento avançados e incentivar a pesquisa competitiva.

Os planejadores são avaliados através de somente três critérios: número de problemas resolvidos, tempo total usado, e do comprimento total do plano solução. Critérios que não consideram outros fatores que influenciam no desempenho de um sistema, como por exemplo a estrutura de dados e a linguagem em que está implementado.

Os problemas propostos são especificados em *PDDL* (seção 2.2), e buscam explorar de forma bastante abrangente as características dos planejadores, indo deste problemas simples, resolvidos rapidamente, até problemas complexos resultando na maioria das vezes em explosões combinatoriais.

A primeira competição, realizada em 1998, foi dividida em duas categorias: uma para planejadores que suportavam apenas características descritas pela linguagem *STRIPS* e outra para os que suportam as características descritas pela linguagem *ADL* [Ped89], que estende a primeira para efeitos condicionais e quantificadores.

Essa competição teve a participação dos seguintes planejadores: *BLACKBOX* [KS99], *HSP* [BG98], *IPP* [KNHD97] e *STAN* [LF99]. Estes planejadores foram comparados a partir de duas baterias de problemas: a primeira composta por 140 problemas em 5 domínios diferentes e a segunda com 15 problemas em 3 domínios diferentes, com o objetivo de refinar os resultados da primeira.

Em 2000 aconteceu a segunda competição que já contou com a participação de quinze planejadores. Entre eles estavam o *BLACKBOX*, o *HSP2* (nova versão do HSP) e o *R*, que é uma reimplentação do *STRIPS* original. Os planejadores *STAN*, *HSP2*, *MIPS* e *R*, obtiveram bons resultados, mas o melhor foi o *FF*, proposto por Hoffman e Nebel [HN01], que utiliza o grafo de planos juntamente com busca heurística.

Na competição de 2002 foi apresentado uma nova versão da linguagem *PDDL* denominada *PDDL*2.1. Adicionaram na versão anterior o planejamento com incerteza e temporal, aumentando significativamente a complexidade dos oito domínios envolvidos. Essa competição contou com a participação de quatorze competidores e obteve a vitória geral do planejador *LPG*.

Na última competição, realizada em 2004, foi também apresentada uma nova versão da linguagem *PDDL*, agora nomeada *PDDL*2.2, que possibilitou a descrição de predicados derivados e eventos condicionais iniciais. Contou com a participação de vinte competidores resolvendo problemas de sete domínios. Novamente foi referenciada a dificuldade em se declarar um ganhador fazendo-se necessário um agrupamento dos planejadores segundo seus desempenhos. Além disso os organizadores enfatizaram para que não fossem analisados somente a classificação por eles definida, mas que fossem observados os resultados obtidos pelos planejadores juntamente com a descrição dos domínios e as técnicas utilizadas por eles. O *Fast Downward*, desenvolvido por Malte Helmert e Silvia Richter, foi o planejador citado como vencedor no planejamento clássico [EHLY04].

Ainda nessa competição foi reafirmada a dificuldade em se comparar os planejadores somente considerando o tempo total gasto para resolver os problemas pois desconsideraram o tempo de leitura e análise sintática da linguagem *PDDL* em que os problemas estavam descritos.

Mas mesmo assim, as competições ainda classificam seus planejadores considerando apenas os resultados finais dos planejadores envolvendo o tempo gasto e o tamanho do plano, não levando em consideração as diferenças de implementações existentes.

O ambiente *IPE*, a ser apresentado no próximo capítulo, permite um estudo mais completo de planejadores possibilitando construí-los e compará-los mais eficientemente. É um ambiente didático, de fácil entendimento, que permite o desenvolvimento em grupo de maneira facilitada. Todos os planejadores seguem a mesma estrutura de implementação, diferenciando-se somente pelo algoritmo e estrutura de representação empregados. Assim, teremos um ambiente de desenvolvimento que supre as necessidades até aqui apresentadas.

# Capítulo 3

# Ambiente de Planejamento Ipê

Nesse capítulo apresentaremos o *IPE - Ipê Planning Enviromment* (Ambiente de Planejamento Ipê) que é um ambiente para desenvolvimento de planejadores possibilitando implementar diferentes algoritmos e estruturas de simplificação utilizando a mesma linguagem de programação e as mesmas características estruturais.

Devido ao grande número de planejadores atualmente encontrados na literatura e a diversidade de suas implementações é difícil compará-los eficientemente e muito menos estudá-los. A Competição de Planejadores [Bac00] tem esse objetivo mas não consegue cumprir totalmente seu papel pois existem fatores, como linguagens e estruturas de implementação, que influenciam no desempenho de um sistema e não são considerados.

Nas seções seguintes apresentaremos a arquitetura *IPE*, apresentando a estrutura de classes definida e implementada e depois exemplificando o seu uso a partir de três planejadores já implementados.

## 3.1   A arquitetura *IPE*

Antes de apresentar a estrutura e o funcionamento do ambiente *IPE* em mais detalhes é importante apresentarmos uma análise detalhada sobre o funcionamento de um planejador.

O processo de solução de um problema de planejamento inicia a partir de arquivos contendo a descrição de um problema a ser resolvido (por exemplo arquivos em *PDDL*). Um *Analisador Sintático* verifica a corretude da descrição e codifica essas informações em uma estrutura interna desejada, objetivando facilitar a indexação dessas informações. Com base nas informações do problema de planejamento indexadas, o *Codificador* gera uma nova estrutura com um melhor poder de representação destas informações (por exemplo: grafo de planos, instância *SAT* ou uma Rede de Petri). Finalmente esta representação é utilizada por algum *Resolvedor* que busca a solução para o problema.

Como pudemos observar um planejador manipula as informações de um problema de planejamento através de diversas estruturas. A figura 3.1 apresenta o fluxo de informações entre estas estruturas, desde a definição de um problema até o plano solução propriamente dito. Os círculos representam as estruturas de informação e os retângulos representam os processos responsáveis pela codificação de uma estrutura em outra.

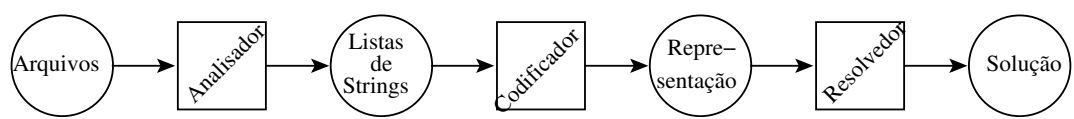

Figura 3.1: Fluxo das informações manipuladas para a solução de um problema de planejamento.

A partir do fluxo de informações, elaboramos um diagrama de classe segundo o paradigma de programação orientada à objeto, que nos permite uma definição mais detalhada dos componentes formadores de um planejador. A figura 3.2 apresenta o diagrama de classes simplificado do *IPE*, uma vez que apresentamos somente os principais atributos e métodos de cada classe. Esse diagrama também refere-se à implementação efetuada nesse trabalho e apresenta ainda possíveis extensões do ambiente *IPE*.

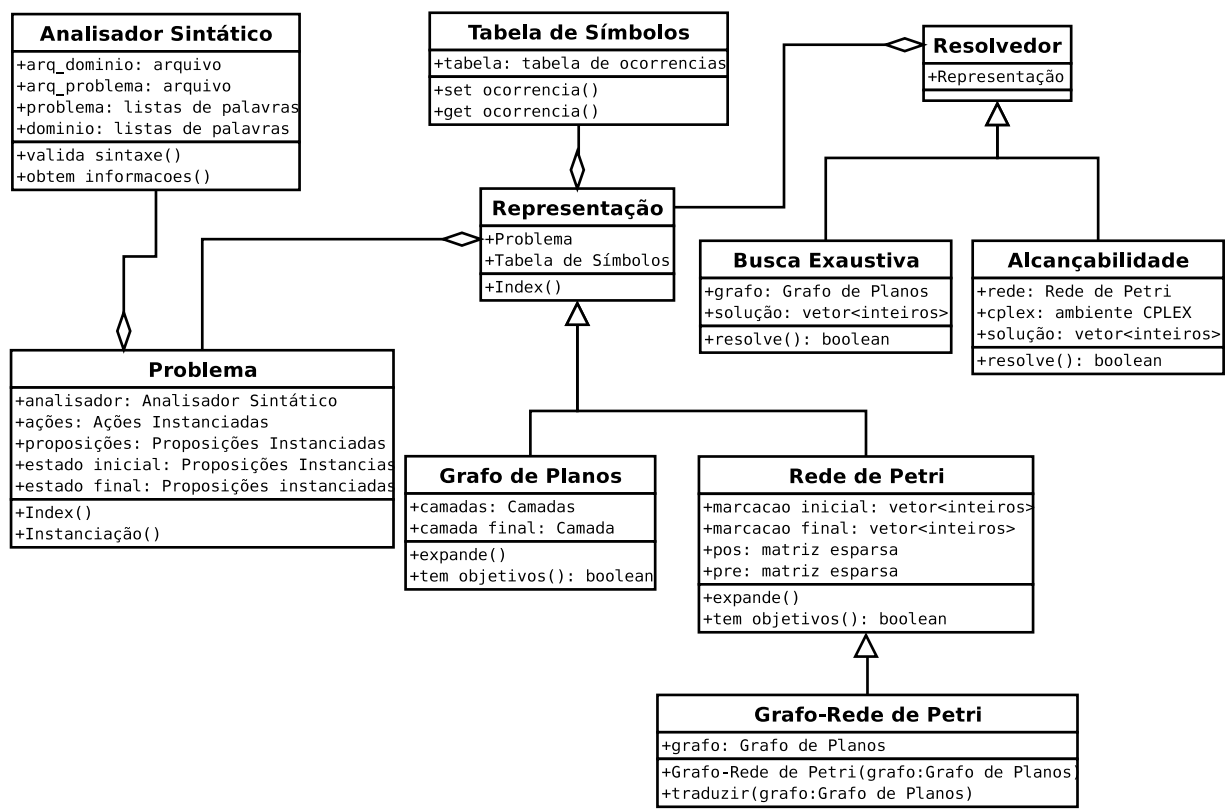

Figura 3.2: Diagrama de classes simplificado do *IPE*.

Seguindo a idéia apresentada anteriormente, as principais classes do sistema são o *Resolvedor*, a *Representação* e o *Analisador Sintático*.

A classe *Analisador Sintático* é formada por métodos que fazem a análise sintática do arquivo de descrição do domínio e do problema, disponibilizando seu conteúdo através de uma estrutura de acesso mais rápida, as listas de palavras. Esta classe é implementada para reconhecimento e validação da linguagem *PDDL*, apresentada na seção 2.2, pois atualmente é a linguagem padrão para a definição dos problemas de planejamento. É implementada em C++ e Bison.

A classe *Problema* apresenta as informações referentes ao domínio e a um problema deste domínio. Ela inclui uma classe *Analisador Sintático*, classe que disponibiliza as informações referentes ao problema. A classe *Problema* faz o processo de instanciação tanto de ações quanto de proposições e apresenta métodos responsáveis pelo processo de indexação dessas informações, aqui representado pelo método *Index()*, que fornece o índice de cada ação ou proposição instanciada. Apresenta ainda o estado inicial e final, formados por proposições instanciadas.

Na classe *Representação* é construída a classe *Tabela de Símbolos*, que auxilia no processo de busca das informações, sendo que cada representação tem a sua própria tabela. Também apresenta métodos de acesso a ocorrências, índices e à tradução destes nas respectivas informações.

A partir da classe *Representação*, o processo se dá apenas através dos índices obtidos através da classe *Problema*. Esta classe é um estereótipo, pois nela somente está definida a estrutura comum de qualquer representação. As classes "filhas" implementarão seus tipos de dados referentes a sua própria representação. Por exemplo, a classe *Grafo de Planos*, *Rede de Petri* e *Grafo-Rede de Petri*, já implementadas. O ambiente possibilita a criação de novas representações. Todas estas classes serão representações que implementarão características distintas.

A classe *Tabela de Símbolos* é uma estrutura de acesso direto implementada a partir de listas de índices e vetores de ocorrências desses índices. Na *Tabela de Símbolos* disponível pela classe *Representação* os índices são referentes as ações e proposições, mas é possível estender os índices para características necessárias em cada representação, como é o caso da classe *Rede de Petri* que insere os índices referentes as transições e lugares da rede. Esta classe possibilita acesso direto tanto à informação quanto às ocorrências dessas informações em toda a estrutura da classe *Representação*.

A classe *Resolvedor* contém uma representação utilizada na aplicação de seu algoritmo de resolução. A representação ainda é responsável pela conversão dos índices em suas respectivas informações não-numéricas que serão utilizadas para apresentar a solução para o usuário do sistema. Esta classe também é um estereótipo, nela somente é definida a estrutura comum de qualquer resolvedor. As classes "filhas" implementarão seus algoritmos sobre sua própria classe *Representação*. Por exemplo, a classe *Busca Exaustiva* que aplica seu algoritmo de busca exaustiva na classe *Grafo de Planos*.

Um planejador é construído pela união destas classes a partir da classe *Resolvedor*. Basicamente um planejador resolverá o problema de planejamento a partir do seu algoritmo aplicado em uma representação que armazena as informações do problema e do domínio.

A presente versão do *IPE* é desenvolvida em C++ para a plataforma GNU/Linux usando o paradigma de programação orientado à objeto pois permite um desenvolvimento independente dos componentes formadores de um planejador. É um projeto desenvolvido pela equipe de planejamento do Laboratório de Inteligência Computacional da Universidade Federal o Paraná - UFPR.

A estrutura projetada nos possibilita uma grande variação de combinações entre resolvedores e representações. Permite a implementação de antigos algoritmos nas novas estruturas de representação existentes. Assim, temos uma plataforma de desenvolvimento flexível e didática que permite implementar diferentes planejadores a partir da mesma estrutura, possibilitando que estes sejam comparados mais efetivamente pois somente apresentarão diferença no algoritmo de busca e/ou representação utilizada.

## 3.2   Construindo planejadores

Como vimos na figura 3.1, independente da representação ou do resolvedor que se deseja desenvolver, tem-se a necessidade de resolver dois problemas iniciais: a leitura e validação dos arquivos *PDDL*, que contêm a descrição do problema a ser resolvido e a instanciação das ações e predicados. Ambos já estão implementados no *IPE* sendo necessário apenas utilizá-los.

Primeiramente deve-se criar um objeto da classe *Analisador Sintático* que faz a análise sintática dos arquivos de domínio e problema escritos em *PDDL*, e a disponibilização de métodos de acesso destas informações. Essa classe necessita de dois parâmetros: o arquivo do domínio e o arquivo do problema.

O segundo objeto a ser criado, responsável pela instanciação das ações e predicados de um referido problema, deve ser uma instância da classe *Problema*. Esse objeto necessita de um parâmetro, o objeto *Analisador Sintático* criado anteriormente. Quando esse objeto é criado, torna-se disponível um conjunto de métodos de acesso a todas proposições e ações instanciadas.

Os passos seguintes são: criar a representação, que armazena as informações do problema de forma organizada para facilitar a solução; e o resolvedor, que encontra uma solução para o problema utilizando a representação. Esses passos são distintos para cada planejador, assim são apresentados a seguir três planejadores que estão implementados no *IPE*.

### 3.2.1   *GRAPHPLAN*-1

Esse planejador foi implementado baseando-se no algoritmo do *GRAPHPLAN*, descrito na seção 2.4.1. Ele se utiliza da representação grafo de planos, apresentada na seção 2.3.2, na qual aplica um algoritmo de busca exaustiva para encontrar o plano solução.

Tendo um objeto da classe *Problema* devidamente instanciado, o próximo passo é criar um objeto da classe *Representação*. No caso do *GRAPHPLAN* a representação utilizada é o grafo de planos. Assim deve-se desenvolver uma sub-classe da classe *Representação*, denominada *Grafo de Planos*, capaz de construir e armazenar o grafo de planos.

A classe *Grafo de Planos*, implementada no *IPE*, possui um construtor que recebe como parâmetro um objeto da classe *Problema*. A partir desse objeto é gerado o estado inicial do grafo de planos e o estado final desejado. Para a expansão do grafo, a classe *Grafo de Planos* apresenta o método expande. Esse método adiciona uma nova camada de ações válidas no grafo e adiciona uma nova camada de proposições que são os efeitos das ações aplicadas.

Um outro método necessário para a criação de um grafo de planos, é o método tem_objetivos. Esse método é o critério de parada para a expansão do grafo, pois verifica se a última camada do grafo gerado até o momento contém todas as proposições do estado final desejado sem conflitos.

Assim, ao instanciar um objeto da classe *Grafo de Planos*, é possível criar um grafo de planos completamente, bastando executar o método expande até que o método tem_objetivos retorne verdadeiro. Esse processo gera o grafo de planos com uma solução possível.

Para extrair o plano, tarefa realizada por um resolvedor que utilize o grafo de planos, é necessário criar um objeto que recebe o grafo de planos e faz uma busca pelo plano solução nesse grafo. O objeto criado no *IPE* é uma instância da classe *Busca Exaustiva*, que, por sua vez, é uma sub-classe da classe *Resolvedor*.

A classe *Busca Exaustiva* recebe como parâmetro um objeto da classe *Representação*, que no caso do *GRAPHPLAN* é da classe *Grafo de Planos*. Essa classe disponibiliza o método *resolve* que faz uma busca exaustiva no grafo de planos para encontrar o plano solução para o problema. Esse método retorna positivo se encontrar um plano e falso se não encontrar, sendo assim necessário fazer uma nova expansão do grafo e, posteriormente, uma nova busca. Esse processo ocorre até que o plano solução seja encontrado ou até que o processo pare por limitações de tempo ou memória.

Tendo todo esse cenário construído, basta agora construir o código principal, mostrado na figura 3.3. Na linha 1 os arquivos que contêm o domínio e o problema são processados. Na linha 2 cria-se a representação interna do problema e domínio. Na linha 3 o grafo de planos é criado. Na linha 4 o grafo de planos é expandido até que a última camada contenha o estado final sem conflitos. Na linha 5 o *GRAPHPLAN*-1 é criado como um objeto *Busca Exaustiva* que recebe como parâmetro o *Grafo de Planos*. Na linha 6 tenta-se encontrar uma solução. Se não foi possível encontrar a solução, uma nova expansão do grafo é realizada.

```
main()
{
1    Analisador_Sintatico analisador("dominio.pddl", "problema.pddl");
2    Problema problema(&analisador);
3    Grafo_de_Planos grafo(&problema);
4    while ( ! grafo.tem_objetivos() )
            grafo.expande();
5    Busca_Exaustiva graphplan-1(&grafo);
6    while ( ! graphplan-1.resolve() )
            grafo.expande();
}
```

Figura 3.3: Código principal do *GRAPHPLAN*-1

### 3.2.2 *PETRIPLAN*-1

O algoritmo do *PETRIPLAN*-1 foi descrito na seção 2.4.2. Esse planejador se utiliza da representação rede de Petri gerada a partir de um grafo de planos e aplica um algoritmo de alcançabilidade de sub-marcação utilizando-se de PI para encontrar um plano solução.

Como foi citado anteriormente, tem-se um objeto da classe *Problema* devidamente instanciado, assim deve-se criar um objeto da classe *Grafo de Planos*, descrita na seção anterior. Tendo o objeto *Grafo de Planos* devidamente instanciado, construir o grafo, a partir do método **expande**, até que o método **tem_objetivos** retorne verdadeiro, o que indica que o grafo gerado contém todas as proposições do estado final desejado sem conflitos.

O próximo passo é criar um objeto capaz de traduzir o grafo de planos em uma Rede de Petri. Esse objeto é uma instância da classe *Grafo-Rede de Petri*, que é construída a partir da classe *Rede de Petri*, que por sua vez é uma sub-classe da classe *Representação*. A classe *Rede de Petri* é aqui utilizada como estrutura para armazenar uma Rede de Petri representada por matrizes. A classe *Grafo-Rede de Petri* possui um construtor que recebe como parâmetro um objeto da classe *Grafo de Planos* e o método *traduzir* que traduz o grafo de planos na Rede de Petri, ou seja, lê o grafo armazenado no objeto *Grafo de Planos* e o traduz para a estrutura da classe *Rede de Petri*.

Para extrair o plano, tarefa realizada por um resolvedor que utilize a Rede de Petri, é necessário criar um objeto que recebe a Rede de Petri e implemente a alcançabilidade de sub-marcação. O objeto criado no *IPE* é uma instância da classe *Alcançabilidade*, que, por sua vez, é uma sub-classe da classe *Resolvedor*.

A classe *Alcançabilidade* recebe como parâmetro um objeto da classe *Representação*, que no caso do *PETRIPLAN*-1 é da classe *Rede de Petri*. Essa classe disponibiliza o

```
main()
{
1    Analisador_Sintatico analisador("dominio.pddl", "problema.pddl");
2    Problema problema(&analisador);
3    Grafo_de_Planos grafo(&problema);
4    while ( ! grafo.tem_objetivos() )
          grafo.expande();
5    Grafo-Rede_de_Petri rede(&grafo);
6    Alcancabilidade petriplan-1(&rede);
7    while ( ! petriplan-1.resolve() )
     {
          grafo.expande();
          rede.traduzir(&grafo);
     }
}
```

Figura 3.4: Código principal do *PETRIPLAN*-1

método *resolve* que encontra o plano solução para o problema resolvendo o problema de *PI* gerado a partir da Rede de Petri. Observe que não é necessário usar *PI* para resolver a alcançabilidade na Rede de Petri, podendo ser substituído por uma busca exaustiva por exemplo.

Esse método retorna positivo se encontrar um plano, e falso se não encontrar, sendo assim necessário fazer uma nova expansão do grafo, uma nova tradução para Rede e posteriormente, uma nova tentativa de resolução. Esse processo ocorre até que o plano solução seja encontrado, ou que o processo pare por limitações de tempo ou memória.

Tendo todo esse cenário construído, basta agora construir o código principal, mostrado na figura 3.4. Na linha 1 os arquivos que contêm o domínio e o problema são processados. Na linha 2 cria-se a representação interna do problema e domínio. Na linha 3 o grafo de planos é criado. Na linha 4 o grafo de planos é expandido até que a última camada contenha o estado final sem conflitos. Na linha 5 a *Rede de Petri* é criada a partir do grafo de planos. Na linha 6 o *PETRIPLAN*-1 é criado como um objeto *Alcançabilidade* que recebe como parâmetro a *Rede de Petri*. Na linha 7 tenta-se encontrar uma solução. Se não é encontrada, o grafo de planos é novamente expandido e a rede de Petri é novamente obtida.

### 3.2.3 *PETRIPLAN*-2

O algoritmo de *PETRIPLAN*-2 é uma reimplementação do *PETRIPLAN*-1 que faz a construção de uma rede de Petri diretamente dos arquivos *PDDL*. A construção da rede de Petri é feita seguindo o processo de construção do grafo de planos e assim aplicando as regras de tradução para a rede.

Como apresentado anteriormente, a classe *Rede de Petri* é responsável pelo armazenamento da rede de Petri. Essa classe possui um construtor que recebe como parâmetro um objeto da classe *Problema*. Foi adicionado o processo de construção da rede, segundo descrito acima. Esse processo é implementado através do método expande. Cada vez que esse método é executado, é adicionado uma "camada" na rede, de modo semelhante ao que é feito no grafo de planos.

Também foi adicionado na classe *Rede de Petri* o método tem_objetivos. Esse método tem a mesma funcionalidade apresentada na classe *Grafo de Planos*, verificar se todas as proposições do estado final desejado estão presentes e sem conflitos, mas essa informação agora é obtida a partir da estrutura da rede de Petri.

Tendo um objeto da classe *Problema* devidamente instanciado, cria-se um objeto da classe *Rede de Petri*. Tendo o objeto *Rede de Petri* devidamente instanciado, basta construir a rede, a partir do método expande, até que o método tem_objetivos retorne verdadeiro.

Assim como no *PETRIPLAN-1*, é necessário agora extrair o plano a partir de uma rede de Petri, tarefa realizada pela classe *Alcançabilidade*. Como já foi visto na seção anterior, tendo um objeto dessa classe, basta executar o método *resolve* até que este retorne positivo, fazendo novas expansões caso contrário.

Tendo todo esse cenário construído, basta agora construir o código principal, mostrado na figura 3.5. Na linha 1 os arquivos que contêm o domínio e o problema são processados. Na linha 2 cria-se a representação interna do problema e domínio. Na linha 3 a rede de Petri é criada. Na linha 4 a rede de Petri é expandida até que a última camada contenha o estado final sem conflitos. Na linha 5 o *PETRIPLAN-2* é criado como um objeto *Alcançabilidade* que recebe como parâmetro a *Rede de Petri*. Na linha 6 tenta-se encontrar uma solução. Se não é encontrada, a rede de Petri é novamente expandida.

```
main()
{
1    Analisador_Sintatico analisador("dominio.pddl", "problema.pddl");
2    Problema problema(&analisador);
3    Rede_de_Petri rede(&problema);
4    while ( ! rede.tem_objetivos() )
         rede.expande();
5    Alcancabilidade petriplan-2(&rede);
6    while ( ! petriplan-2.resolve() )
     {
         rede.expande();
     }
}
```

Figura 3.5: Código principal do *PETRIPLAN-2*.

No próximo capítulo será apresentado uma análise dos três planejadores aqui descritos e implementados no *IPE* de forma a justificar e identificar seu potencial.

# Capítulo 4

# Experimentos

Neste capítulo são mostrados alguns resultados obtidos utilizando o ambiente proposto. O objetivo aqui é mostrar informações mais relevantes do que apenas o tempo final gasto permitindo uma análise mais detalhada dos planejadores.

Os experimentos foram realizados em uma máquina com um processador AMD de 2 Ghz, 512 MB de memória RAM e o sistema operacional Debian GNU/Linux.

Os planejadores aqui analisados foram os mesmos descritos na seção anterior: *GRAPH-PLAN*1, *PETRIPLAN*1 e o *PETRIPLAN*2 pois todos estão completamente implementados no *IPE*.

Por motivo de espaço, foi limitado a análise dos planejadores na solução de três problemas de domínios conhecidos: mundo de blocos, robô (*gripper*) e logística; mas a atual implementação suporta a resolução de outros domínios e problemas do planejamento clássico.

As complexidades dos problemas escolhidos foram definidas a partir de execuções de diferentes níveis, sendo que os problemas aqui apresentados possibilitam uma melhor análise por não serem de níveis intermediários. Assim poderemos apresentar mais detalhadamente as particularidades dos planejadores ao resolverem os problemas.

O primeiro problema é do clássico domínio "mundo de blocos". O estado inicial e final estão representados pela figura 4.1. Existe apenas uma ação que permite mover um bloco de um lugar para outro, sendo que somente podem ser movidos os blocos livres, ou seja, que não possuem nenhum outro bloco sobre ele.

O segundo problema, do domínio *gripper*, é constituído por um robô com duas garras, duas salas e algumas bolas. Ele pode carregar uma bola em cada garra. O objetivo do problema aqui utilizado é levar quatro bolas de uma sala para outra. Existem três ações possíveis: pegar uma bola com uma garra estando em uma sala, soltar a bola da garra em uma sala e mover de uma sala para outra.

O terceiro problema, do domínio de logística, consiste em distribuir seis pacotes em duas cidades, sendo que cada cidade possui um depósito e um aeroporto, ambos capazes de armazenar os pacotes. Para isso são disponibilizados dois caminhões, um em cada cidade, que podem se locomover apenas dentro da cidade, e um avião que pode voar entre os aeroportos transportando os pacotes entre as cidades.

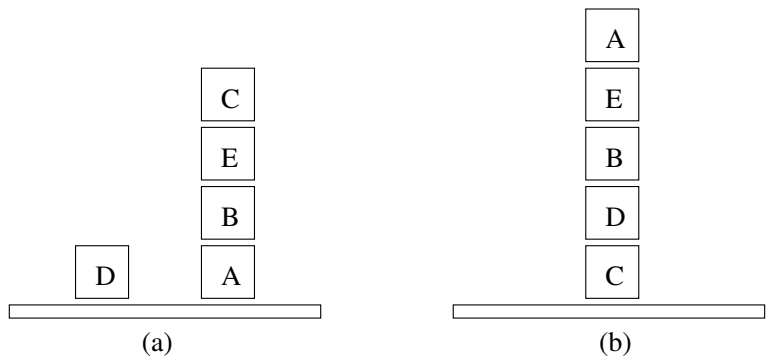

Figura 4.1: Estado inicial (a) e estado final do problema com cinco blocos do domínio mundo de blocos.

Para uma melhor análise os resultados estão separados ainda pelas características do processo de resolução e estrutura utilizada. Assim, são apresentados inicialmente os resultados obtidos pelo *GRAPHPLAN*1 e *PETRIPLAN*1 pois utilizam a mesma representação, diferenciando-se nos processos seguintes pois utilizam resolvedores diferentes.

O primeiro problema a ser utilizado na análise é o problema dos blocos. Observando a tabela 4.1 pode-se identificar a grande diferença no tempo de execução entre os dois planejadores, o *GRAPHPLAN* executou 24 vezes mais rápido que o *PETRIPLAN*1, informação apresentada na primeira linha. O número de ações e o tamanho do plano indicam a igualdade das soluções encontradas. O tamanho do grafo de planos, indicado pelo número de nodos, arestas e conflitos, demonstra a igualdade também na representação final obtida e utilizada para a busca da solução.

Tabela 4.1: Resultados obtidos pelos planejadores *GRAPHPLAN*1 e *PETRIPLAN*1 no problema dos blocos. O tempo total é apresentado em segundos e o tamanho do plano se refere ao número de passos necessários para se executar o plano.

|  | *GRAPHPLAN*1 | *PETRIPLAN*1 |
|---|---|---|
| *Tempo Total* | 0.17 | 4.18 |
| *Número de Ações* | 7 | 7 |
| *Tamanho do Plano* | 6 | 6 |
| *Número de Nodos* | 531 | 531 |
| *Número de Arestas* | 1138 | 1138 |
| *Número de Mutexes* | 12760 | 12760 |

Mesmo com todas essas informações é difícil analisar e principalmente verificar o que ocasiona a lentidão apresentada pelo *PETRIPLAN*1. Como no ambiente *IPE* tem-se o domínio de todo o processo, geramos a tabela 4.2 que complementa a tabela anterior e apresenta outras informações que são relevantes para a análise.

Tabela 4.2: Resultados obtidos pelos planejadores *GRAPHPLAN*1 e *PETRIPLAN*1 no problema dos blocos. Tempos gastos, em segundos, em cada processo executado pelos planejadores.

|  | *GRAPHPLAN*1 | *PETRIPLAN*1 |
|---|---|---|
| *Tempo do Analisador* | 0.00 | 0.00 |
| *Tempo da Instanciação* | 0.00 | 0.00 |
| *Tempo do Cálc. Mut.* | 0.01 | 0.01 |
| *Tempo das Expansões* | 0.16 | 0.17 |
| *Tempo das Traduções* | - | 0.10 |
| *Tempo das Buscas* | 0.00 | 3.90 |
| *Tempo Total* | 0.17 | 4.18 |

Observando essa tabela verifica-se que os únicos processos responsáveis pelo aumento do tempo são as traduções do grafo de planos para Rede de Petri e as buscas efetuadas, sendo este último o maior responsável. Todos os outros tempos são, desconsiderando o arredondamento, idênticos pois são a mesma implementação. Com todas essas informações pode-se concluir com certeza que o resolvedor utilizado pelo *PETRIPLAN*1 é ineficiente com relação ao tempo, mas gera soluções idênticas às obtidas pelo *GRAPHPLAN*1.

Como foi citado anteriormente, foram agrupados os resultados obtidos para cada problema de acordo com os planejadores, assim, a próxima análise compreenderá os resultados obtidos pelo *PETRIPLAN*1 e *PETRIPLAN*2 na solução do problema dos blocos pois utilizam representações obtidas diferentemente mas um mesmo resolvedor.

Observando a tabela 4.3 pode-se verificar também que a diferença no tempo de execução dos planejadores, apresentada na primeira linha, não é tão grande quanto a diferença anterior, 24 vezes, mas significativa. O *PETRIPLAN*2 é 1.6 vezes mais rápido que o *PETRIPLAN*1. O número de ações e o tamanho do plano indicam a igualdade das soluções encontradas. Pode-se verificar a diferença no tamanho da rede de Petri obtida, apresentado pelo número de linhas, colunas, valores diferentes de zero e conflitos. O principal desses itens é os valores diferentes de zero, o *PETRIPLAN*2 apresenta 2.6 vezes menos que o *PETRIPLAN*1. Esse número influencia diretamente na resolução pois estes são os valores que serão utilizados para criar o problema de PI. Essa informação já é uma característica importante para identificar a diferença no tempo, mas não o suficiente.

Ao analisar mais informações, apresentadas na tabela 4.4 que apresenta os tempos gastos por cada processo executado pelo planejador, pode-se observar dois fatores importantes. O primeiro é a diferença no tempo gasto pelo resolvedor, que é justificada pelo tamanho da rede reduzido apresentado anteriormente. O segundo é a diferença relativamente grande no tempo gasto pelo processo de expansão, no qual o *PETRIPLAN*2 gasta $4,29$ vezes mais que o *PETRIPLAN*1. Esse valor foi obtido dividindo-se o tempo gasto nas expansões do $PETRIPLAN2$ pelo $PETRIPLAN1$. Essa diferença é justificada pela estrutura de implementação da matriz utilizada para armazenar a rede de Petri.

Tabela 4.3: Resultados obtidos pelos planejadores $PETRIPLAN1$ e $PETRIPLAN2$ no problema dos blocos. O tempo total é apresentado em segundos e o tamanho da Rede de Petri é o número de colunas, linhas e elementos com valores diferentes de zero. Também apresenta a quantidade de conflitos representada na rede.

|  | $PETRIPLAN1$ | $PETRIPLAN2$ |
|---|---|---|
| *Tempo Total* | 4.18 | 2.58 |
| *Número de Ações* | 7 | 7 |
| *Tamanho do Plano* | 6 | 6 |
| *Número de Linhas* | 5111 | 2146 |
| *Número de Colunas* | 427 | 412 |
| *Número de Val. Não Zeros* | 9557 | 3651 |
| *Número de Conflitos* | 4468 | 1505 |

Tabela 4.4: Resultados obtidos pelos planejadores $PETRIPLAN1$ e $PETRIPLAN2$ no problema do mundo de blocos. Tempos gastos, em segundos, em cada processo executado pelos planejadores.

|  | $PETRIPLAN1$ | $PETRIPLAN2$ |
|---|---|---|
| *Tempo do Analisador* | 0.00 | 0.00 |
| *Tempo da Instanciação* | 0.00 | 0.00 |
| *Tempo do Cálc. Mut.* | 0.01 | – |
| *Tempo das Expansões* | 0.17 | 0.73 |
| *Tempo das Traduções* | 0.10 | – |
| *Tempo das Buscas* | 3.90 | 1.84 |
| *Tempo Total* | 4.18 | 2.58 |

A próxima análise é feita sobre o problema do robô. A tabela 4.5 apresenta os resultados obtidos pelo $GRAPHPLAN1$ e $PETRIPLAN1$. Pode-se observar que para esse problema também ocorre uma grande diferença no tempo de execução dos planejadores, informação presente na primeira linha. E as outras informações demonstram a igualdade da solução e representação.

Observando a tabela 4.6 pode-se verificar e confirmar os resultados do problema anterior onde verifica-se que o tempo gasto pelo $PETRIPLAN1$ é quase que totalmente causado pela busca (7.09 segundos) e não pelo processo de tradução (0.16 segundos). Todos os outros tempos são, desconsiderando o arredondamento, idênticos pois são a mesma implementação.

Tabela 4.5: Resultados obtidos pelos planejadores $GRAPHPLAN1$ e $PETRIPLAN1$ para o problema do robô. O tempo total é apresentado em segundos e o tamanho do plano se refere ao número de passos necessários para se executar o plano.

|  | $GRAPHPLAN1$ | $PETRIPLAN1$ |
|---|---|---|
| *Tempo Total* | 0.17 | 7.39 |
| *Número de Ações* | 11 | 11 |
| *Tamanho do Plano* | 7 | 7 |
| *Número de Nodos* | 573 | 573 |
| *Número de Arestas* | 1266 | 1266 |
| *Número de Mutexes* | 12032 | 12032 |

Tabela 4.6: Resultados obtidos pelos planejadores $GRAPHPLAN1$ e $PETRIPLAN1$ para o problema do robô. Tempos gastos, em segundos, em cada processo executado pelos planejadores.

|  | $GRAPHPLAN1$ | $PETRIPLAN1$ |
|---|---|---|
| *Tempo do Analisador* | 0.00 | 0.00 |
| *Tempo da Instanciação* | 0.00 | 0.00 |
| *Tempo do Cálc. Mut.* | 0.00 | 0.00 |
| *Tempo das Expansões* | 0.15 | 0.14 |
| *Tempo das Traduções* | – | 0.16 |
| *Tempo das Buscas* | 0.02 | 7.09 |
| *Tempo Total* | 0.17 | 7.39 |

Com todas essas observações pode-se concluir que o resolvedor utilizado pelo $PETRI$-$PLAN1$ é ineficiente com relação ao tempo, mas gera soluções corretas e idênticas às obtidas pelo $GRAPHPLAN1$.

A próxima análise será feita com base na tabela 4.7 que compreende a resolução do problema do robô pelos planejadores $PETRIPLAN1$ e $PETRIPLAN2$. Observando essa tabela pode-se observar novamente a diferença no tempo de execução dos planejadores, agora não tão intensa, mas significativa.

O número de ações e o tamanho do plano indicam a igualdade das soluções encontradas. Aqui também apresenta-se a diferença no tamanho da rede de Petri obtida, apresentado pelo número de linhas, colunas, valores diferentes de zero e conflitos, informação importante para identificar a diferença no tempo, mas não o suficiente.

Ao analisar mais informações, apresentadas na tabela 4.8 que apresenta os tempos gastos por cada processo executado pelo planejador, pode-se observar dois fatores importantes. O primeiro é a diferença no tempo gasto pelo resolvedor, que é justificada pelo tamanho da rede reduzido apresentado anteriormente. O segundo é que os tempos gastos pelo processo de expansão são quase que equivalentes, bem diferente do que foi obtido no problema anterior onde o $PETRIPLAN2$ foi 4,29 vezes mais lento que o $PETRIPLAN1$

Tabela 4.7: Resultados obtidos pelos planejadores *PETRIPLAN*1 e *PETRIPLAN*2 para o problema do robô. O tempo total é apresentado em segundos e o tamanho da Rede de Petri é o número de colunas, linhas e elementos com valores diferentes de zero. Também apresenta a quantidade de conflitos representada na rede.

|  | *PETRIPLAN*1 | *PETRIPLAN*2 |
|---|---|---|
| *Tempo Total* | 7.39 | 3.53 |
| *Número de Ações* | 11 | 11 |
| *Tamanho do Plano* | 7 | 7 |
| *Número de Linhas* | 5312 | 2374 |
| *Número de Colunas* | 454 | 454 |
| *Número de Val. Não Zeros* | 9906 | 4030 |
| *Número de Conflitos* | 4595 | 1656 |

no processo de expansão.

Tabela 4.8: Resultados obtidos pelos planejadores *PETRIPLAN*1 e *PETRIPLAN*2 no problema do robô. Tempos gastos, em segundos, em cada processo executado pelos planejadores.

|  | *PETRIPLAN*1 | *PETRIPLAN*2 |
|---|---|---|
| *Tempo do Analisador* | 0.00 | 0.00 |
| *Tempo da Instanciação* | 0.00 | 0.00 |
| *Tempo do Cálc. Mut.* | 0.00 | – |
| *Tempo das Expansões* | 0.14 | 0.15 |
| *Tempo das Traduções* | 0.16 | – |
| *Tempo das Buscas* | 7.09 | 3.36 |
| *Tempo Total* | 7.39 | 3.52 |

Como têm-se total conhecimento e domínio de todo o processo e estrutura empregados no *IPE* é fácil fazer uma análise mais detalhada para verificar o motivo dessa diferença. Para isso foram obtidas mais informações do processo de expansão do planejador *PETRIPLAN*2 que são apresentados na tabela 4.9.

Tabela 4.9: Resultados obtidos pelo planejador *PETRIPLAN*2 no problema dos blocos e do robô. Tempos gastos, em segundos, em cada expansão executada pelo planejador.

|  | *blocos* | *gripper* |
|---|---|---|
| *Tempo das Expansões*1 | 0.00 | 0.01 |
| *Tempo das Expansões*2 | 0.00 | 0.02 |
| *Tempo das Expansões*3 | 0.03 | 0.05 |
| *Tempo das Expansões*4 | 0.16 | 0.07 |
| *Tempo das Expansões*5 | 0.23 | 0.00 |
| *Tempo das Expansões*6 | 0.30 | 0.00 |
| *Tempo das Expansões*7 | – | 0.00 |
| *Tempo das Expansões* | 0.72 | 0.15 |

Com essas informações pode-se verificar que o *PETRIPLAN*2 no problema dos blocos executa seis expansões e sempre aumenta o tempo gasto em cada expansão. Já no problema do robô o *PETRIPLAN*2 executa uma expansão a mais, sete expansões, mas a partir da quinta o tempo gasto não chega nem a um centésimo.

Essa diferença ocorre devido a um processo de cópia que ocorre no método expande da classe *Rede de Petri*. Quando o grafo de planos está estagnado, ou seja, não existe nenhuma diferença entre as duas últimas camadas, torna-se possível a obtenção da próxima camada a partir de uma cópia da camada anterior. Como a rede de Petri é armazenada em matrizes, quando ocorre a estagnação, é feito apenas uma cópia das linhas e colunas que formam a última camada. Verifica-se então que no problema dos blocos não acontece a estagnação antes que a solução seja possível.

O último experimento realizado compreende os resultados obtidos na resolução do problema de logística. A tabela 4.10 apresenta os resultados obtidos pelo *GRAPHPLAN*1 e *PETRIPLAN*1. O comportamento desses planejadores é equivalente ao retratado nas análises anteriores, podendo-se observar que também apresenta uma grande diferença no tempo de execução dos planejadores mas que as outras informações demonstram a igualdade da solução e representação.

Tabela 4.10: Resultados obtidos pelos planejadores *GRAPHPLAN*1 e *PETRIPLAN*1 para o problema de logística. O tempo total é apresentado em segundos e o tamanho do plano se refere ao número de passos necessários para se executar o plano.

|  | *GRAPHPLAN*1 | *PETRIPLAN*1 |
|---|---|---|
| *Tempo Total* | 0.42 | 11.12 |
| *Número de Ações* | 23 | 23 |
| *Tamanho do Plano* | 9 | 9 |
| *Número de Nodos* | 1297 | 1297 |
| *Número de Arestas* | 2308 | 2308 |
| *Número de Mutexes* | 19230 | 19230 |

Um fato interessante acontece na resolução do problema de logística pelos planejadores *PETRIPLAN*1 e *PETRIPLAN*2, informações apresentadas na tabela 4.11. Observando essa tabela pode-se identificar novamente a diferença no tempo de execução dos planejadores e no tamanho da representação, mas pode-se ainda identificar mais um item importante e que diferencia-se das análises anteriores, item este que é a diferença do número de ações dos planos encontrados.

A diferença no número de ações é um fator importante para avaliação de quão otimizado é o plano. Um plano é mais otimizado quando resolve o problema com o menor número de ações possíveis. Existem ainda planos que solucionam o problema mas apresentam ações que não são necessárias e que não interferem na solução do problema.

Como no ambiente *IPE* ambos planejadores utilizam os mesmos processos e estru-turas iniciais e os mesmos resolvedores, a diferença somente pode ocorrer na estrutura de representação. O *PETRIPLAN*2 gera a rede de Petri a partir da definição do problema considerando os processos de construção do grafo de planos mas não implementa os conflitos recorrentes, já o *PETRIPLAN*1 gera a rede de Petri a partir do grafo de planos.

Tabela 4.11: Resultados obtidos pelos planejadores *PETRIPLAN*1 e *PETRIPLAN*2 para o problema de logística. O tempo total é apresentado em segundos e o tamanho da Rede de Petri é o número de colunas, linhas e elementos com valores diferentes de zero. Também apresenta a quantidade de conflitos representada na rede.

|  | *PETRIPLAN*1 | *PETRIPLAN*2 |
|---|---|---|
| *Tempo Total* | 11.12 | 12.85 |
| *Número de Ações* | 23 | 21 |
| *Tamanho do Plano* | 9 | 9 |
| *Número de Linhas* | 8840 | 3626 |
| *Número de Colunas* | 1048 | 1144 |
| *Número de Val. Não Zeros* | 16172 | 5600 |
| *Número de Conflitos* | 7347 | 1974 |

Com essa informação pode-se verificar que a representação da rede de Petri além de não necessitar que seja implementado os conflitos recorrentes pois representa-os em sua própria estrutura fornece uma solução mais otimizada, uma vez que o *PETRIPLAN*2 apresentou a solução com menor número de ações desnecessárias que o *PETRIPLAN*1 e o *GRAPHPLAN*1. Mesmo assim verificou-se que ambas as soluções ainda contêm ações que são desnecessárias para o problema, por exemplo no domínio de logística os planos encontrados apresentam ações que movimentam o avião da cidade 1 para a cidade 2 e depois da cidade 2 para a cidade 1 enquanto que o caminhão dirige-se para o depósito e é carregado.

Com essas informações pode-se, além de entender melhor o funcionamento de cada planejador, propor e implementar melhorias para os problemas encontrados.

Como pode-se observar, o *IPE* possibilita a construção e análise eficiente de planejadores, permitindo verificar detalhes da execução e estrutura, além de ser uma boa alternativa como ferramenta de auxílio para cursos sobre planejamento. Além da possibilidade de utilizar os processos e estruturas já existentes, o *IPE* permite verificar onde realmente está a diferença dos planejadores, se na estrutura de representação ou no algoritmo empregado, possibilitando ainda um melhor entendimento para possíveis melhoras e inovações.

# Capítulo 5

# Conclusão

Planejadores são sistemas complexos. Para analisá-los e estudá-los é preciso entender seus componentes: leitura do problema em uma linguagem de descrição, representação em uma estrutura de simplificação e algoritmo de resolução. As características de implementação desses componentes influenciam diretamente o desempenho dos planejadores. Assim, comparar planejadores não é tarefa simples, pois cada um possui uma implementação diferente e os métodos encontrados na literatura são ineficientes para analisar e comparar detalhadamente os planejadores.

Neste contexto apresentamos o *IPE*, que é um ambiente que permite o estudo mais detalhado dos planejadores, possibilitando construí-los e compará-los com maior profundidade. Nele é possível implementar os componentes do planejador em uma estrutura equivalente permitindo análises detalhadas de suas particularidades. No desenvolvimento de um novo planejador podemos utilizar os componentes já implementados, por exemplo o analisador sintático e/ou a representação, e desenvolver somente o que é de interesse, por exemplo o algoritmo resolvedor. Assim o *IPE* permite um desenvolvimento rápido e ainda possibilita verificar onde realmente está a diferença dos planejadores, se na estrutura de representação ou no algoritmo empregado, permitindo uma análise mais profunda para possíveis melhoras e inovações.

Estas características tornam o *IPE* uma boa alternativa como ferramenta de auxílio para cursos sobre planejamento, além de facilitar os trabalhos em equipe. Por exemplo, atualmente já existem várias implementações terminadas ou em andamento no *IPE*, além dos três referenciados no capítulo 3. O *AgPlan* [CLPS04] está totalmente implementado no *IPE* e possui uma equipe implementando sua paralelização. Existem mais duas implementações de planejadores que traduzem a rede de Petri em SAT aplicando resolvedores diferenciados. Como nossa análise indicou claramente que o ponto falho do *PETRIPLAN* é o resolvedor, alguns trabalhos no Laboratório de Inteligência Computacional (LIC)[1] foram direcionados para melhor tratar o problema de alcançabilidade em redes de Petri

---

[1]http://www.inf.ufpr.br/lic

[Ben04, Mon04].

Outro importante trabalho no *IPE* é a implementação de uma nova abordagem na tradução do problema de planejamento em uma rede de Petri aproveitando melhor sua capacidade representativa [Sil03]. Esse trabalho foi o motivador para participarmos da IPC - *International Planning Competition* - de 2004, mas não obtivemos sucesso na implementação, em tempo hábil, de um resolvedor mais eficiente para o problema de alcançabilidade em redes de Petri, o que continua em desenvolvimento [CGL$^+$04].

O analisador sintático necessita ser expandido para possibilitar a manipulação de problemas do planejamento não-clássico, que já está em andamento com o auxílio de alunos de uma disciplina da pós-graduação. Outros trabalhos visam ainda a expansão do *IPE* para tratar problemas envolvendo tempo ou recursos, usando como base os algoritmos *PETRIPLAN* implementados [Nov04, Mon04].

Todas as figuras apresentadas nesse trabalho e que representam grafos de plano foram geradas utilizando uma classe implementada no *IPE*. Essa classe também necessita ser expandida para desenhar outras representações além do grafo de planos.

O *IPE* está disponível através do CVS, serviço de desenvolvimento compartilhado, bastando enviar e-mail para ipe@inf.ufpr.br.

# Referências Bibliográficas